\pdfoutput=1
\documentclass[letterpaper, 10 pt, conference]{ieeeconf}
\IEEEoverridecommandlockouts
\usepackage{graphicx}
\usepackage{caption}
\usepackage{subcaption}
\usepackage{amsmath}
\usepackage{amssymb}
\usepackage{bm}
\usepackage{hyperref}
\usepackage{gensymb}
\usepackage{changes}
\definechangesauthor{BS}

\title{\LARGE \bf
Computational Tactile Flow for Anthropomorphic Grippers
}

\author{Kanishka Ganguly$^{1,+}$, Behzad Sadrfaridpour$^{1,+}$, Cornelia Ferm{\"u}ller$^{1}$, Yiannis Aloimonos$^{1}$% <-this % stops a space
\thanks{$^{1}$ Institute for Advanced Computer Studies, University of Maryland, College Park
        {\tt\small \{kganguly,behzad,fer,yiannis\}@umiacs.umd.edu}}%
\thanks{$^{+}$ The authors claim equal contribution for the work presented in this paper.}%
}

\begin{document}
\maketitle
\thispagestyle{empty}
\pagestyle{empty}

%%%%%%%%%%%%%%%%%%%%%%%%%%%%%%%%%%%%%%%%%%%%%%%%%%%%%%%%%%%%%%%%%%%%%%%%%%%%%%%%
\begin{abstract}
Grasping objects requires tight integration between visual and tactile feedback. However, there is an inherent difference in the scale at which both these input modalities operate. It is thus necessary to be able to analyze tactile feedback in isolation in order to gain information about the surface the end-effector is operating on, such that more fine-grained features may be extracted from the surroundings.
For tactile perception of the robot, inspired by the concept of the tactile flow in humans, we present the \textit{computational tactile flow} to improve the analysis of the tactile feedback in robots using a Shadow Dexterous Hand. 

In the computational tactile flow model, given a sequence of  pressure values from the  tactile sensors, we define a virtual surface for the pressure values and define the tactile flow as the optical flow of this surface. We provide case studies that demonstrate how the computational tactile flow maps reveal information on the direction of motion and 3D structure of the surface, and feedback regarding the action being performed by the robot.
\end{abstract}

%%%%%%%%%%%%%%%%%%%%%%%%%%%%%%%%%%%%%%%%%%%%%%%%%%%%%%%%%%%%%%%%%%%%%%%%%%%%%%%%
\section{INTRODUCTION}\label{sec:introduction}
The concept of tactile flow, as it pertains to human manipulation and sensing, has been studied extensively both from a neurological as well as psychological perspective~\cite{bicchi2008tactileillusions,denunzio2017prosthesisfeedback}.
The part of the brain responsible for receiving and processing tactile feedback is called the somatosensory cortex and is subdivided into three main areas, known as Brodmann's areas 3a, 3b, 1 and 2. It is important to understand how the human brain processes and decodes tactile information in order to replicate it on robotic end-effectors. As our understanding of the human brain has developed, so has the understanding of its shortcomings. This has led to a rise in research on tactile/haptic illusions~\cite{Bicchi2003tactileillusions2,bicchi2008tactileillusions}, in which the misperceptions of the tactile signals received by the brain are leveraged to ``fool'' the mind into perceiving sensations that are not in sync with reality. This allows a better understanding of how certain parts of the brain react to different input signals.

We believe that the tactile flow provides a reach understanding of the frictional forces between surfaces of the hand and the object. The friction interaction can be categorized to static and dynamic interactions. In the static interaction, the gripper does not move relative the object but it performs an action on or with the object such as opening the lid of a jar or pushing and screwing a screw in a wall. In the dynamic interaction, the gripper and the object have a relative motion with respect to each other mostly for detection of the surface.

Most of the existing research on the tactile flow has been performed on human subjects~\cite{bicchi2008tactileillusions,Bicchi2003tactileillusions2,harris2017tactilecues} but has mostly considered the dynamic friction interaction. We use that as an inspiration for some of our case studies and present other case studies to cover the static friction interaction. It is important to attempt to replicate human studies on robotic systems, if we are to create robots that function at a level similar to humans. This is especially true for anthropomorphic grippers, such as the Shadow Dexterous Hand we use, since effective grasping using robotic systems is still far from being a reality. To the best of our knowledge, this is the first paper to simulate tactile flow on an electro-mechanical sensor, the BioTac, and we attempt to replicate some common and representative tactile experiments on the robot. We perform our experiments on specially designed experimental surfaces, which capture a variety of scenarios. We are able to use this computational tactile flow to discover the direction of motion of the finger, detect different surface properties such as width and height of textures, recover 3D information about surface textures and also angles of these textures for the dynamic interaction. Moreover, we are able to detect some of the tactile flow patterns for various manipulation actions.

The organization of the rest of this paper is as follows. Section~\ref{sec:related} provides a summary of the related works. Section~\ref{sec:tactile_flow} explains the details of our approach to computing the tactile flow. Our case studies and their results are presented in Section~\ref{sec:experiments}. A discussion of our findings and ideas for future works are presented in Section~\ref{sec:conclusion}.

%%%%%%%%%%%%%%%%%%%%%%%%%%%%%%%%%%%%%%%%%%%%%%%%%%%%%%%%%%%%%%%%%%%%%%%%%%%%%%%%
\section{Related Work}\label{sec:related}
The concept of tactile flow is not new and has been discussed and studied at length over the years. It has been found that tactile flow in humans is a highly sensitive and important source of proprioception and can override other cues to self-motion~\cite{harris2017tactilecues}. Harris, et al. conclude in this work that due to the sensitivity of tactile flow, they act as an ``emergency override'' and even minimal cues are enough to promote stability in a subject. This demonstrates the importance of having good tactile sensory mechanisms in robotic systems, if they are to become pervasive in day to day applications.

In a pilot study conducted by Ricciardi et al.~\cite{ricciardi2004cortical}, fMRI studies of the human brain have discovered that visual and tactile flow both result in activation of the V5/MT cortex, suggesting that a similar process occurs in the human brain when decoding motion cues from either visual or tactile cues. They define tactile flow as the ``flow associated with displacement of iso-stress curves on the surface of contacting fingertips.`` This quantifies to our ability to perceive relative motion as well as changes in pressure on the surfaces in contact with the fingertip(s). In our presented work, we use this as the basis of our approach and show that conventional optical flow algorithms can indeed be used to discover tactile flow, using the BioTac~\cite{biotac} sensors. We call this process and our output, computational tactile flow using robotic sensors.

On a more practical level, tactile feedback remains crucial to human grasping and manipulation. In Chapter 7.3 of~\cite{humanrobothands}, the authors discuss the crucial nature of tactile slip, which encodes information about the relative motion between the skin and the surface. We demonstrate that our computational tactile flow also encodes this information and can be used as control feedback for dynamic robot grasping. Similarly, applications such as robot-assisted surgery can greatly benefit from tactile feedback, since it results in reduced (and more accurate) grasping forces on objects~\cite{king2009feedbacksurgery} and thus improve control over the robotic system.

We also take note of the concept of ``tactile illusions'' or ``haptic illusions''~\cite{Bicchi2003tactileillusions2,bicchi2008tactileillusions} which leverages the aforementioned property of similar visual and tactile activation regions, in order to induce misrepresentations caused by the dynamic tactile stimulation of the fingertips, and they have been shown to be similar to how optical illusions work and can be explained by tactile flow perception, which is the analog to optical flow. In future work, we would like to consider how haptic illusions affect our computation of simulated tactile flow and how we may find mechanisms to compensate for them.
%%%%%%%%%%%%%%%%%%%%%%%%%%%%%%%%%%%%%%%%%%%%%%%%%%%%%%%%%%%%%%%%%%%%%%%%%%%%%%%%
%\cite{garcia2019tactilegcn}
\section{Computational Tactile Flow}\label{sec:tactile_flow}
%%%%%%%%%%%%%%%%%%%%%%%%%%%%%%%%%%%%%%%%%%%%%%%%%%%%%%%%%%%%%%%%%%%%%%%%%%%%%%%%
%When a robotic hand performs a manipulation task, it applies some haptic forces to the surface of the manipulated object. Based on the manipulation task and the surfaces of the robotic hand and the object, these haptic forces change spatially and temporally. 
%Our goal is to capture these variations, defined as the tactile flow between the robotic finger and the surface of the object being manipulated. We then use the tactile flow to extract information about the surface and, consequently, the task being performed.
As it discussed in Sections~\ref{sec:introduction}~and~\ref{sec:related}, people construct tactile flows for their tactile sensory stimuli.
In this section, we propose a method for computing the tactile flow for a robotic gripper equipped with tactile sensors.
%In the following sections, we explain our method for finding the tactile flow in our setup and demonstrate some experiments and test our results on a custom dataset gathered for this task.
\subsection{Robotic Hardware}
\begin{figure}[tbh!]
\centering
    \includegraphics[width=0.225\textwidth]{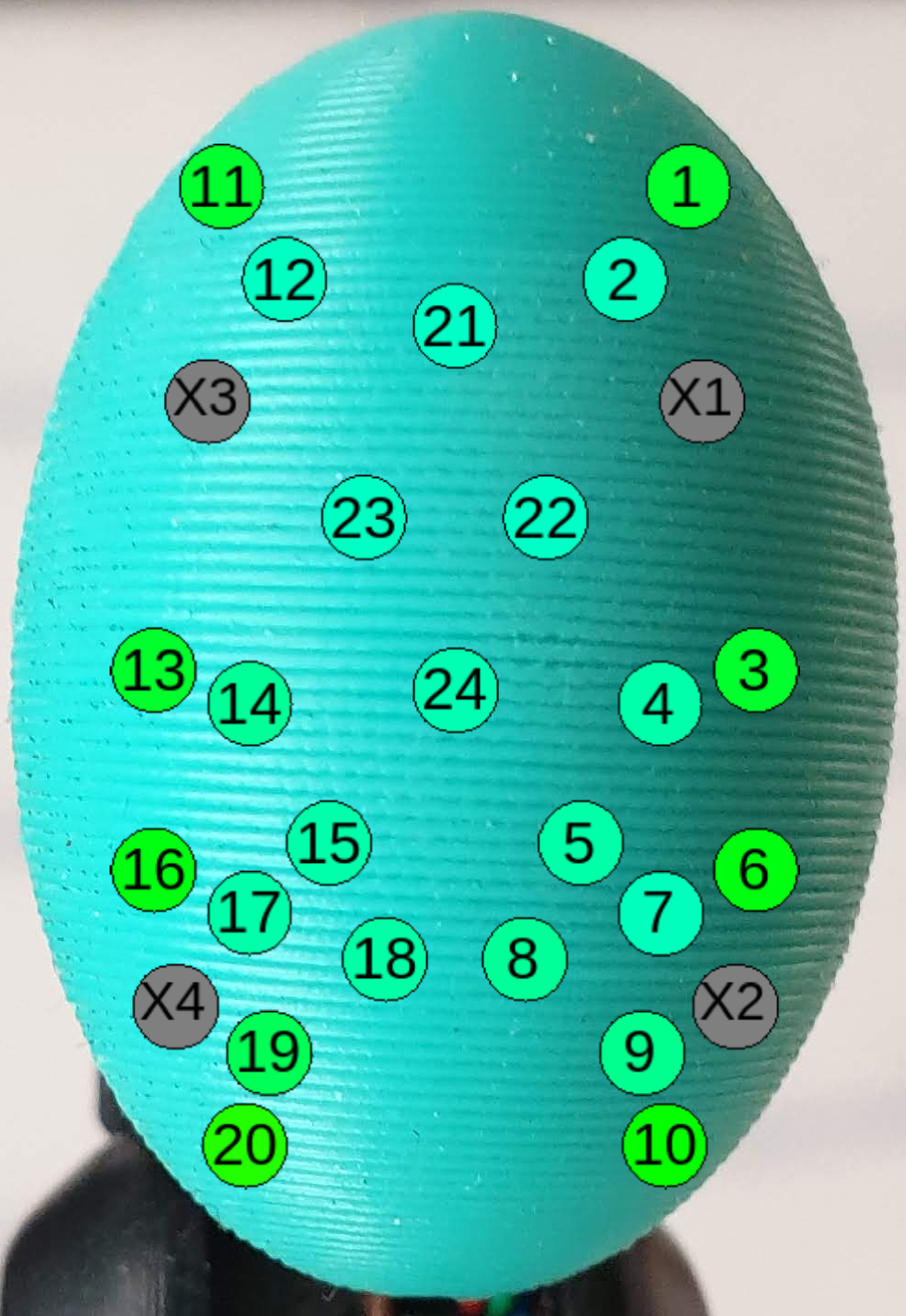}
    \caption{The BioTac SP sensor and layout of the taxels.}
    \label{fig:biotac}
\end{figure}

We are using Shadow Dexterous Hand equipped with BioTac SP~\cite{biotac} tactile sensors on its five fingers. Each of the BioTac sensors has multimodal sensory capabilities. They sense contact forces, micro-vibrations, and heat flux. All of the sensory electronics are attached to a rigid core which receives the sensory information from a soft elastomeric skin through an incompressible conductive fluid. The attached sensors include (i) a hydro-acoustic pressure sensor for measuring the pressure of the whole fluid, (ii) a thermistor for measuring the vibrations and heat flux, and, (iii) 24 distributed electrodes, called taxels (tactile pixels), for measuring the changes of their electrical impedance in response to the deformation of the skin. Note that the impedance values approximate the pressure that the fluid between the skin and core applies at each electrode. Fig.~\ref{fig:biotac} shows the BioTac SP tactile sensor with a layout of the electrode positions. The taxel 3-dimensional positions are not given by the manufacturer but provided in~\cite{garcia2019tactilegcn}. We use the same data in this work.
\subsection{Smoothing the Data}
\begin{figure}[tbh!]
\centering
     \begin{subfigure}{0.225\textwidth}
        \includegraphics[width=\textwidth]{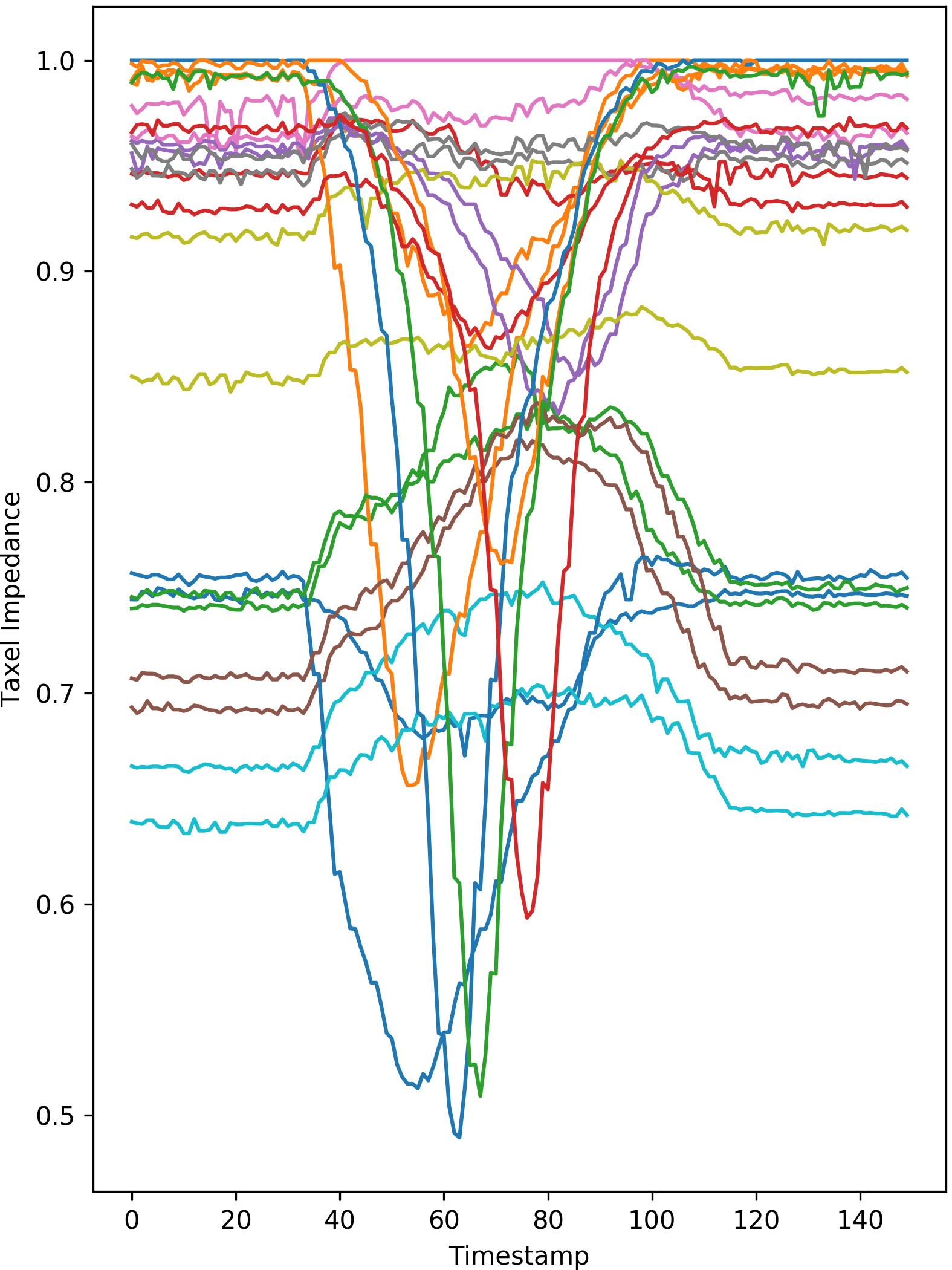}
        \caption{Raw data}\label{fig:measurements}
    \end{subfigure}
    \quad
    \begin{subfigure}{0.225\textwidth}
        \includegraphics[width=\textwidth]{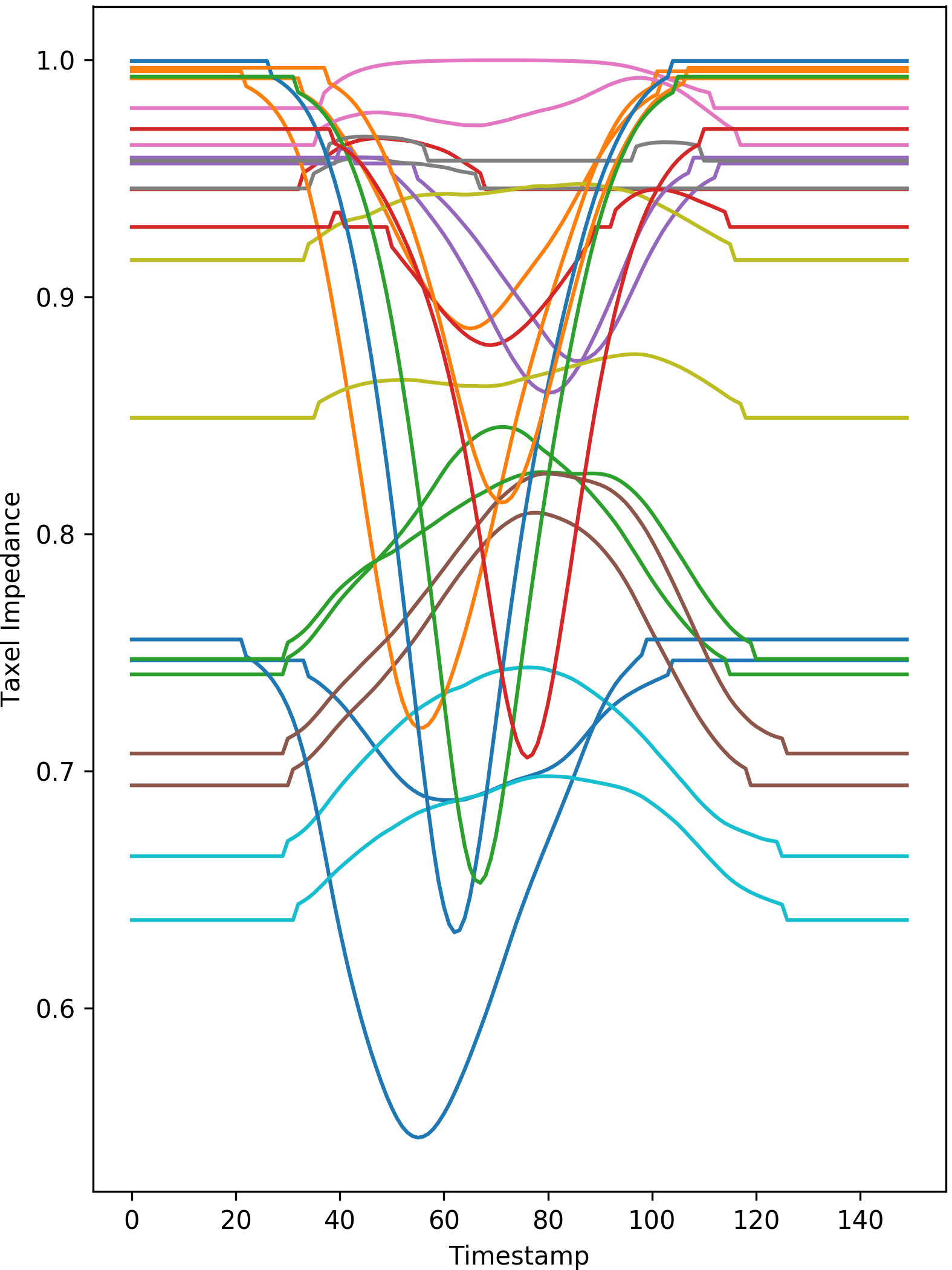}
        \caption{Smoothed data}\label{fig:kalman}
    \end{subfigure}
\caption{Sample trajectories of the sensor values.}
\end{figure}

The measured values of the sensor is noisy as it can be seen in Fig.~\ref{fig:measurements}. Therefore, before performing any interpretation on the data, we smooth the data using Kalman method, as shown in Fig.~\ref{fig:kalman}. First, we apply a Kalman filter with zero velocity model. We define the vector for the sensor readings of the 24 taxels at each time step $t$ as
\begin{equation}
\bm{p}(t)=\begin{bmatrix}p(\bm{x}_1(t)) & \dots & p(\bm{x}_{24}(t))    \end{bmatrix}^T.
\end{equation}
We choose the state transition matrix as $\bm{\Phi}=\bm{I}_{24}$, where $\bm{I}_{24}$ is the $24\times 24$ identity matrix. We choose $\bm{R}=0.005\bm{I}_{24}$ and $\bm{Q}=0.00015\bm{I}_{24}$ as the initial covariances of the measurement and process noises. The filter iteratively updates the data to $\bm{p}(t|t)$ for the measurement $\bm{p}(t)$ as
\begin{eqnarray}
&&\bm{p}(t|t-1) = \bm{\Phi}\bm{p}(t-1|t-1) \nonumber \\
&&\bm{S}(t|t-1) =  \bm{\Phi}\bm{S}(t-1|t-1)\bm{\Phi}^T + \bm{Q} \nonumber \\
&&\bm{K}(t) = \bm{S}(t|t-1)[\bm{S}(t|t-1)+\bm{R}]^{-1} \nonumber \\
&&\bm{S}(t|t) =  (\bm{I}_{24}-\bm{K}(t))\bm{S}(t|t-1) \nonumber \\
&&\bm{p}(t|t) = \bm{p}(t|t-1)+\bm{K}(t)(\bm{p}(t)-\bm{p}(t|t-1)), \nonumber
\end{eqnarray}
where $\bm{S}$ is the state covariance matrix. We record the data over the time period $T = \{ 1  \& \dots \& N_T\}$. After applying the filter, we smooth the data given the whole observed data to $\bm{p}(t|T)$ by backward iterations as
\begin{eqnarray}
\bm{L}(t)&=&\bm{S}(t-1|t-1)\bm{\Phi}^T\bm{S}(t|t-1)^{-1}  \nonumber \\
\bm{S}(t-1|T)&=&\bm{S}(t-1|t-1)+   \nonumber \\
 &\ & \bm{L}(t)(\bm{S}(t|T)-\bm{S}(t|t-1))\bm{L}(t)^T  \nonumber \\
\bm{p}(t-1|T)&=&\bm{p}(t-1|t-1)+ \nonumber \\
 &\ &  \bm{L}(t)(\bm{p}(t|T)-\bm{p}(t|t-1)). \nonumber
\end{eqnarray}
\subsection{Tactile Data Interpolation}
\begin{figure}[bh!]
\centering
    \includegraphics[width=0.4\textwidth]{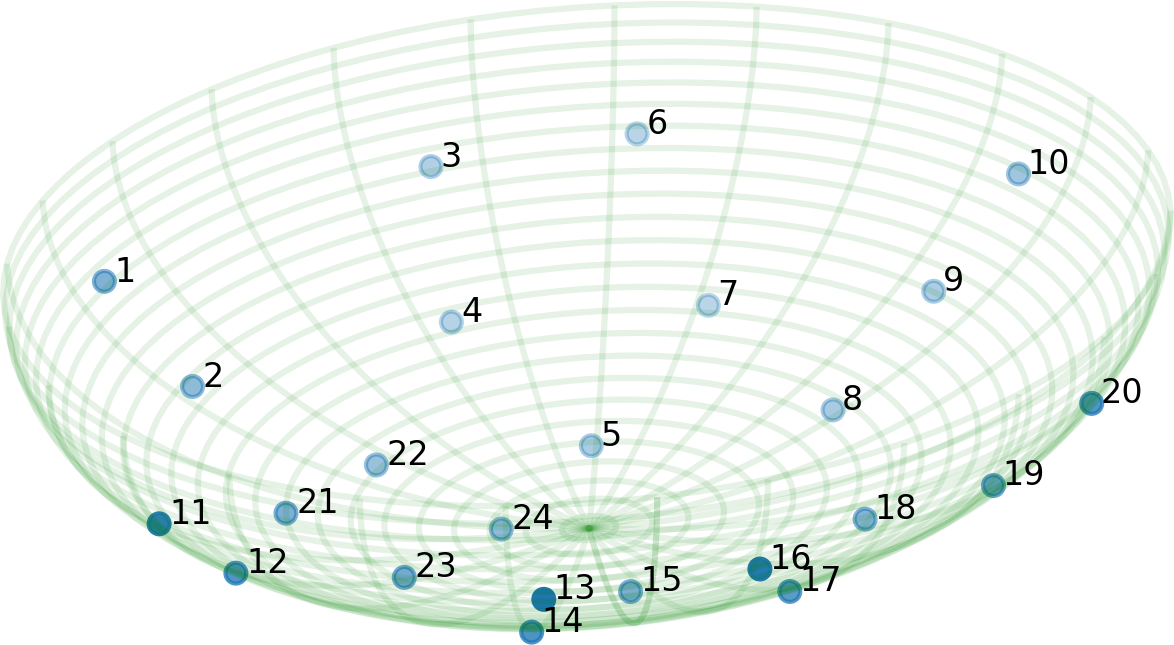}
    \caption{The taxels and the half ellipsoid.}\label{fig:biotac_3d}
\end{figure}

Taxels provide observed information only on their specific locations on the BioTac sensors. To simulate the tactile flow on the sensor, we need the taxel values over the whole surface of the sensor. We can find these values by interpolating the data on the surface of the sensor. The 3D model of the sensor is not given but, using the least square error method, we realized that a half ellipsoid fits the 24 taxel positions well. Fig.~\ref{fig:biotac_3d} shows a 3D visualization of the 24 taxels and the fitted half-ellipsoid.

\begin{figure}[tbh!]
\centering
    \includegraphics[width=0.45\textwidth]{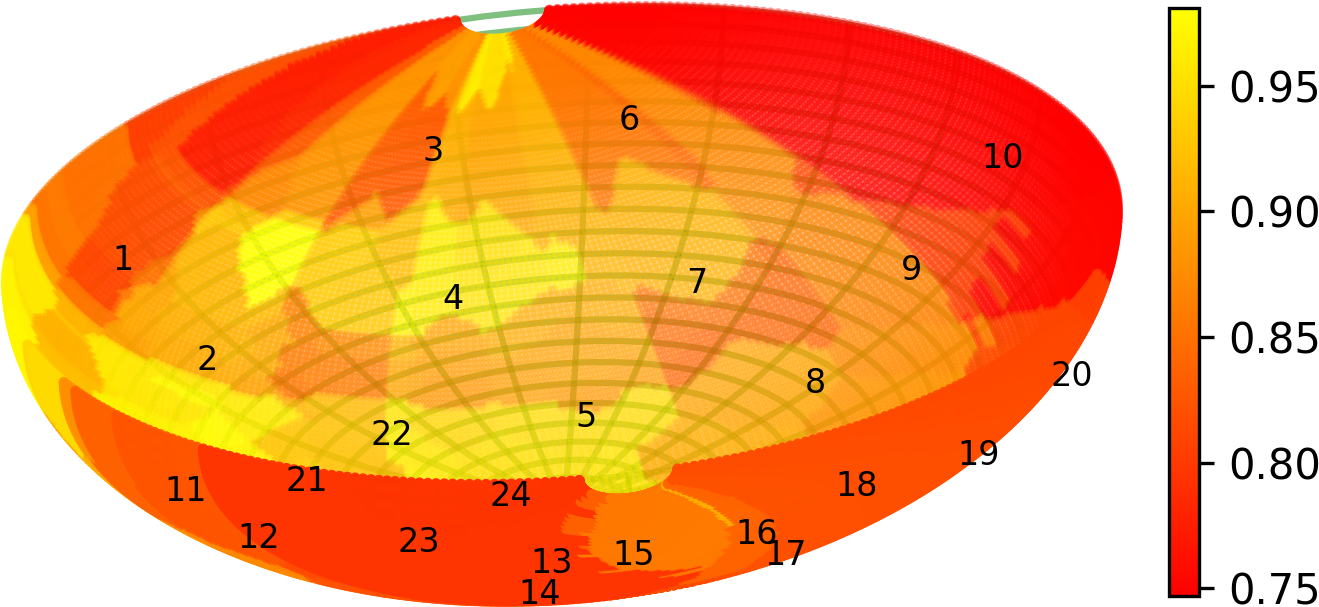}
    \caption{A sample plot of the interpolated impedance values on the half ellipsoid.}\label{fig:sample_inter_ellipsoid}
\end{figure}

For the interpolation of taxel values at $\bm{x},\ \bm{x}_i \in \mathbb{R}^3$ on the core of the sensor, first note that a 3D point $\bm{x}$ on the ellipsoid can be presented by two parameters. The relation between the 3D representation, $\bm{x}$, and its corresponding 2D representation, $\bm{\theta}=[\theta \ \phi]^T$, can be written as
\begin{eqnarray}
&& x = a\sin{\theta}\cos{\phi} \nonumber \\
&& y = b\sin{\theta}\sin{\phi} \nonumber \\
&& z = c\cos{\theta} \label{eqn:ellipsoid}
\end{eqnarray}
where $a$, $b$ and $c$ are the parameters of the half-ellipsoid, and 
\begin{equation}
% && 0 \leq \theta \leq \pi \quad \text{and} \quad \pi \leq \phi \leq 2\pi. \nonumber
\end{equation}
Therefore, there is a mapping $f:[0,\pi]\times[\pi,2\pi]\rightarrow \mathbb{R}^3$ such that $f(\bm{\theta})=\bm{x}$. The interpolation problem of taxel values can be approximated using a Gaussian kernel as 
%For the interpolation of taxel values at $\bm{x},\ \bm{x}_i \in \mathbb{R}^3$ on the core of the sensor, we use a Gaussian kernel as
%To find the tactile flow, we interpolate the data over the whole area of the sensor using Gaussian kernel. We project the taxels on a 2D surface projected beneath the BioTac sensor and call it the \emph{tactile frame}. For the taxel points $\bm{x},\ \bm{x}_i \in \mathbb{R}^2$ on the tactile frame, we consider the kernel as
\begin{equation}\label{eqn:inter3d}
k(\bm{\theta},\bm{\theta}_i)=\exp{\left(-\frac{1}{2\sigma^2}d(\bm{\theta},\bm{\theta}_i)\right)},
\end{equation}
where $d(\bm{\theta},\bm{\theta}_i)$ is the shortest distance between $\bm{x}=f(\bm{\theta})$ and $\bm{x}_i=f(\bm{\theta}_i)$ on the surface of sensor (called geodesic) and $\sigma$ is the kernel parameter. The fitted ellipsoid on the surface is a tri-axial ellipsoid with a parameter for each axis. The problem of finding the minimum distance between two points on a tri-axial ellipsoid does not have a closed form analytical solution. Here, we approximate the geodesic distance numerically by integrating the euclidean distances of a set of consecutive points on the surface which are between $\bm{x}$ and $\bm{x}_i$. The taxel values at each point, $p(\bm{\theta})$ can be estimated as
\begin{eqnarray}
\hat{p}(\bm{\theta})=\frac{\sum_{i=1}^{24}k(\bm{\theta},\bm{\theta}_i)p(\bm{\theta}_i)}{\sum_{i=1}^{24}k(\bm{\theta},\bm{\theta}_i)},
\end{eqnarray}
 where $p(\bm{\theta}_i)$ is the measured impedance value of the $i$-th electrode and $\hat{p}(\bm{\theta})$ is the estimated value of $p(\bm{\theta})$. Fig.~\ref{fig:sample_inter_ellipsoid} shows a sample of the contour plot of the interpolation of the impedance values.
% Fig.~\ref{fig:sample_frame} shows a tactile frame and its contour plot for the free-touch condition of the BioTac sensor without any external forces. The contour plot distinguishes the range of the impedance values in five regions.
% %
% \begin{figure}[tbh!]
% \centering
%      \begin{subfigure}{0.225\textwidth}
%         \includegraphics[width=\textwidth]{img/fig_interpolation.png}
%         \caption{Tactile frame}\label{fig:interpolation}
%     \end{subfigure}
%     \quad
%     \begin{subfigure}{0.225\textwidth}
%         \includegraphics[width=\textwidth]{img/fig_contour.png}
%         \caption{Taxel values}\label{fig:contour}
%     \end{subfigure}
%     \caption{A sample tactile frame and its contour plot.}\label{fig:sample_frame}
% \end{figure}
%

\begin{figure}[tbh!]
\centering
    \includegraphics[width=.45\textwidth]{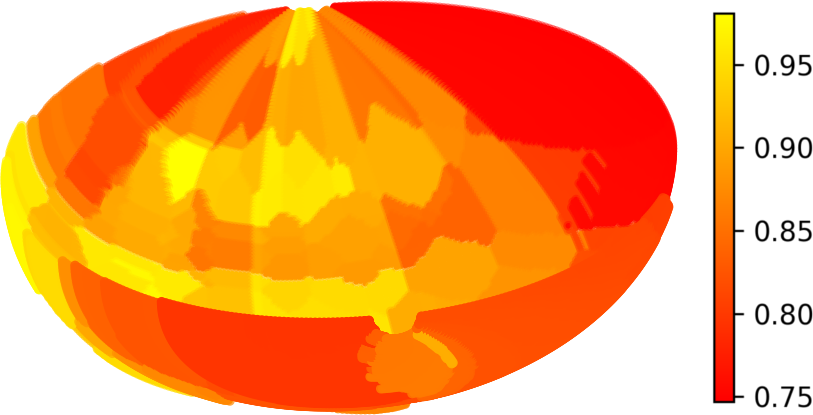}
    \caption{The tactile surface.}\label{fig:3Dsurface}
\end{figure}
\begin{figure}[tbh!]
\centering
     \begin{subfigure}{0.225\textwidth}
        \includegraphics[width=\textwidth]{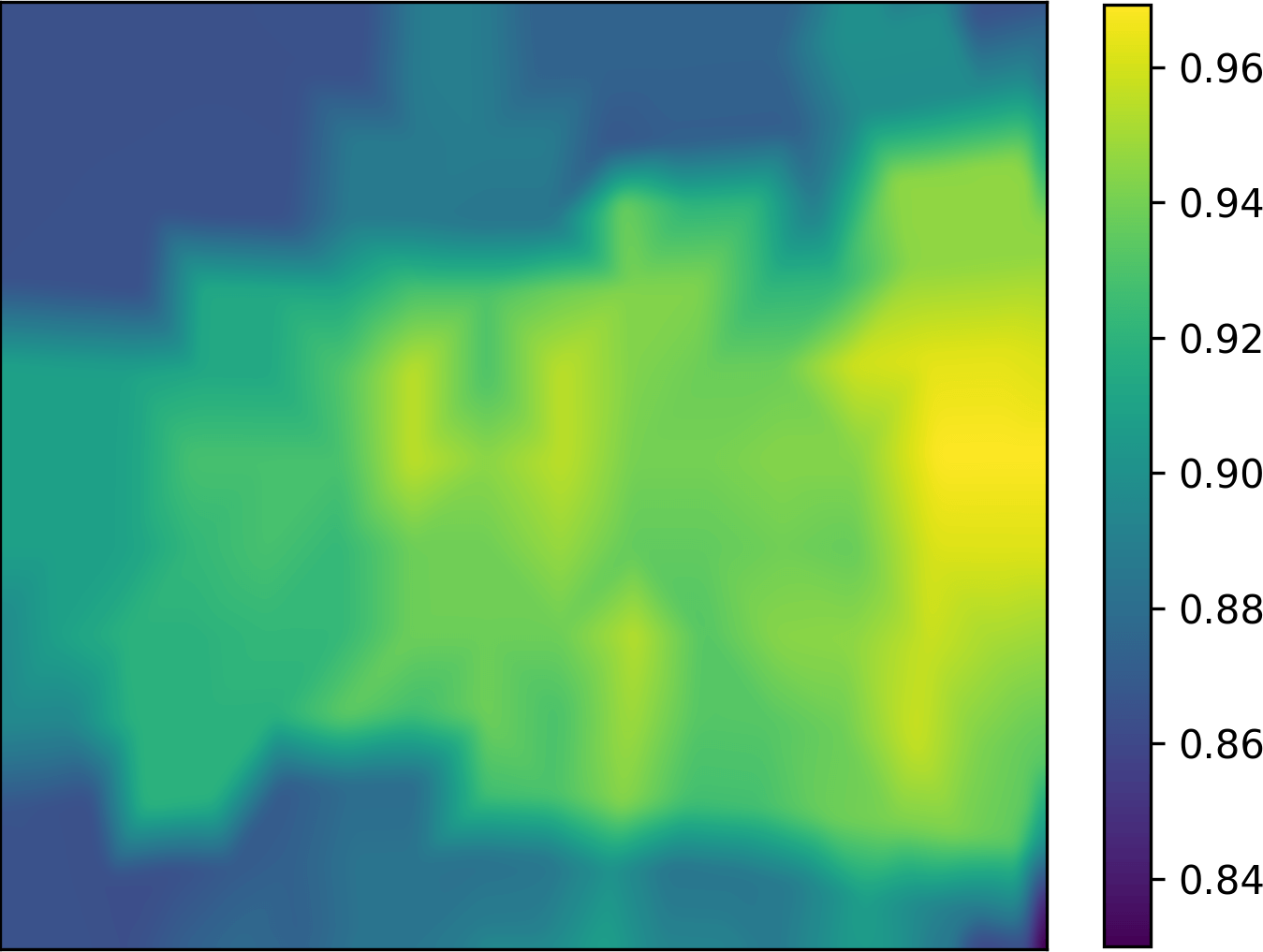}
        \caption{Tactile frame}\label{fig:sample_frame}
    \end{subfigure}
    \quad
    \begin{subfigure}{0.225\textwidth}
        \includegraphics[width=\textwidth]{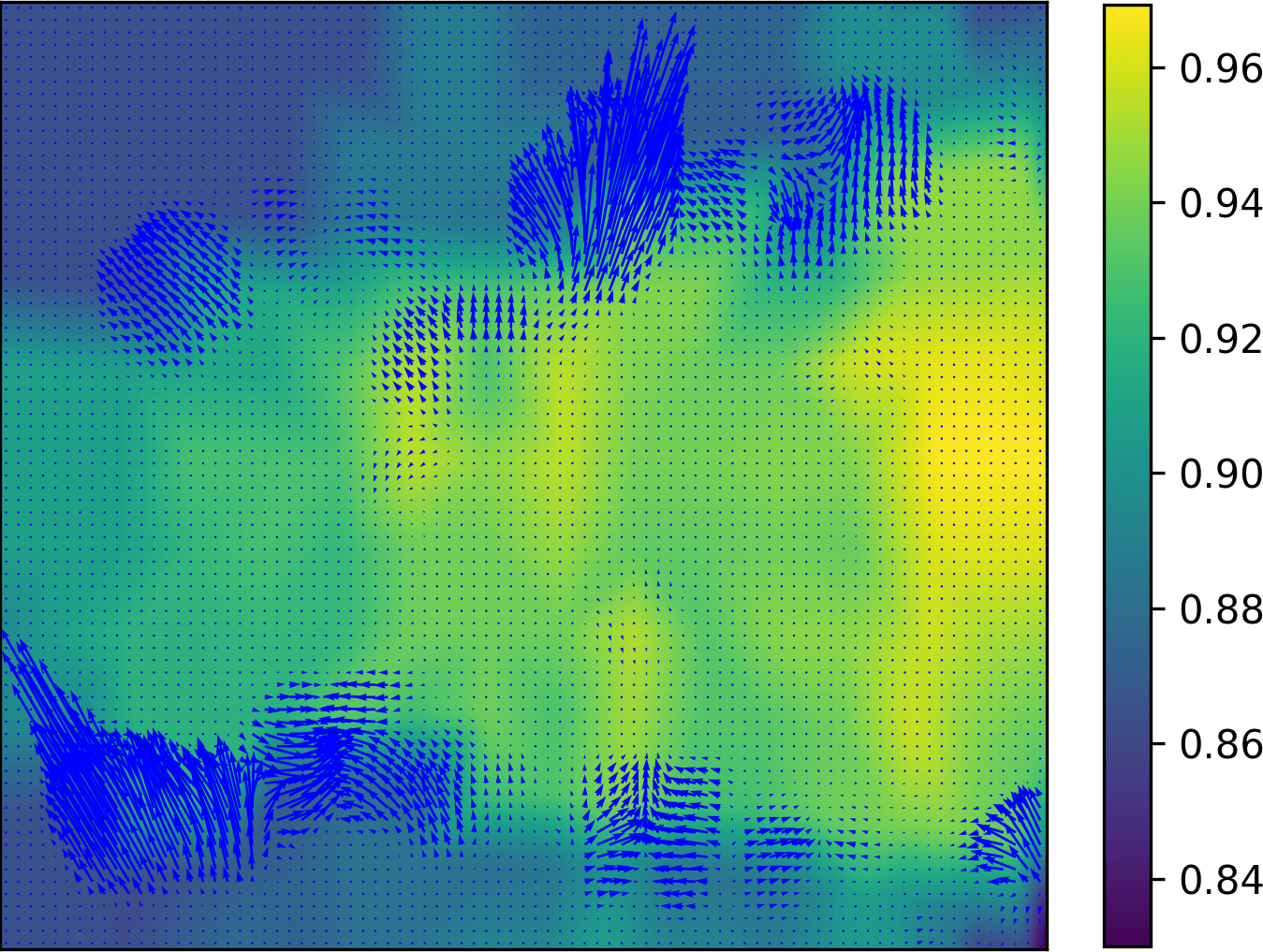}
        \caption{Tactile flow}\label{fig:sample_flow}
    \end{subfigure}
    \caption{A sample tactile frame of projection in $x-y$ plane and its corresponding tactile flow.}
    \label{fig:sample_flow_proj}
\end{figure}
\begin{figure*}[tbh!]
\centering
     \begin{subfigure}{0.22\textwidth}
        \includegraphics[width=\textwidth]{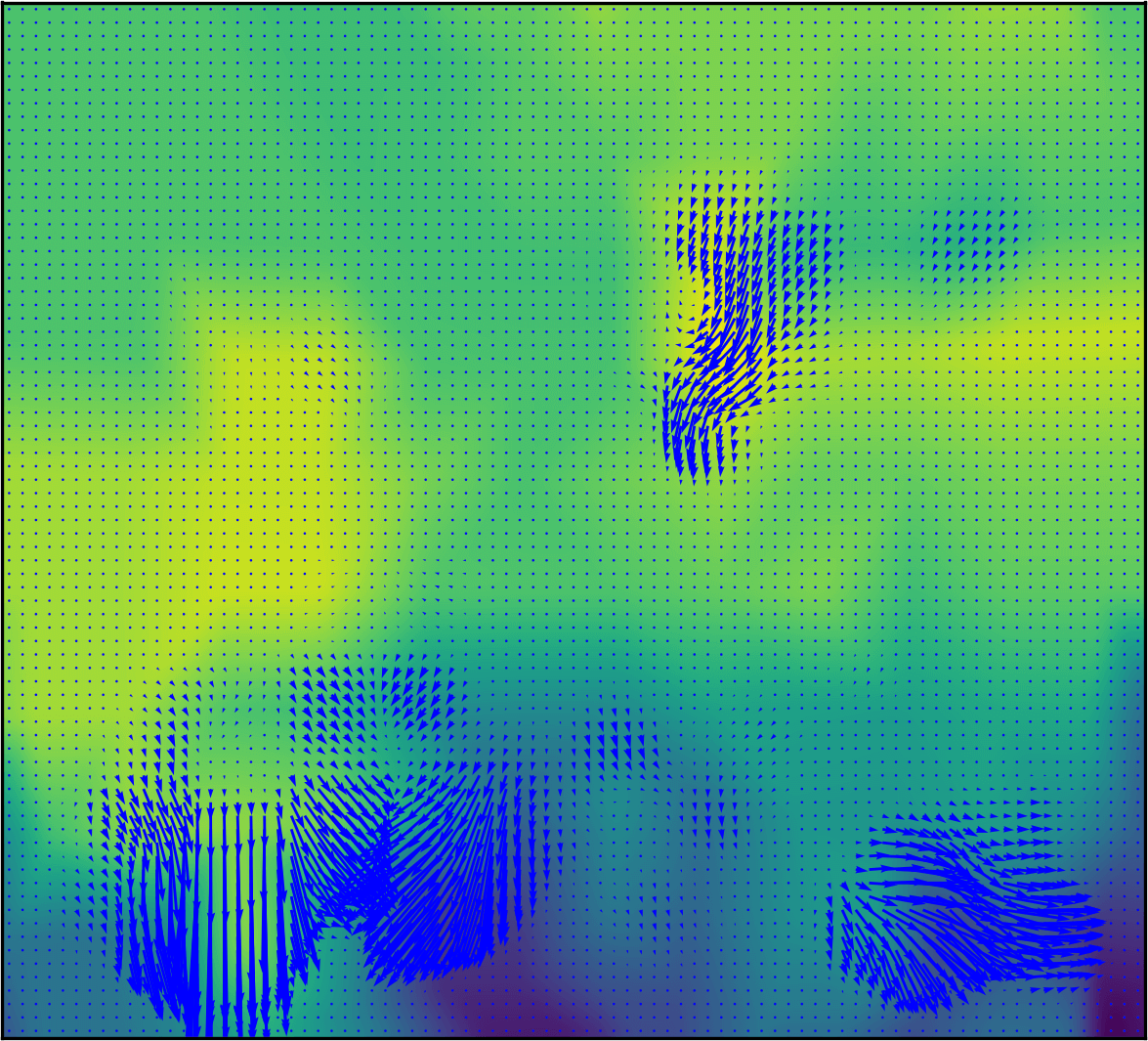}
    \end{subfigure}
    \
    \begin{subfigure}{0.22\textwidth}
        \includegraphics[width=\textwidth]{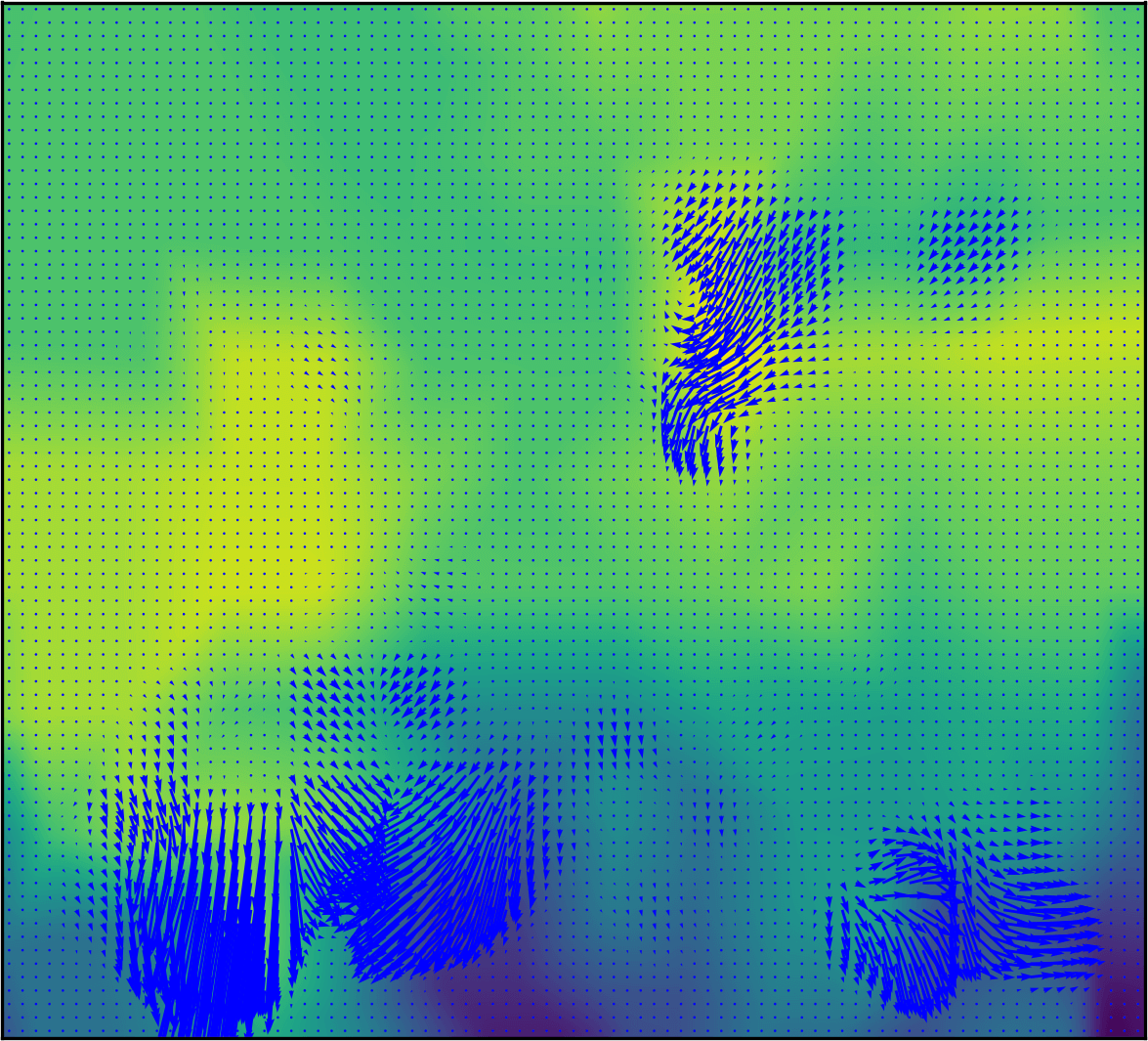}
    \end{subfigure}
    \
    \begin{subfigure}{0.22\textwidth}
        \includegraphics[width=\textwidth]{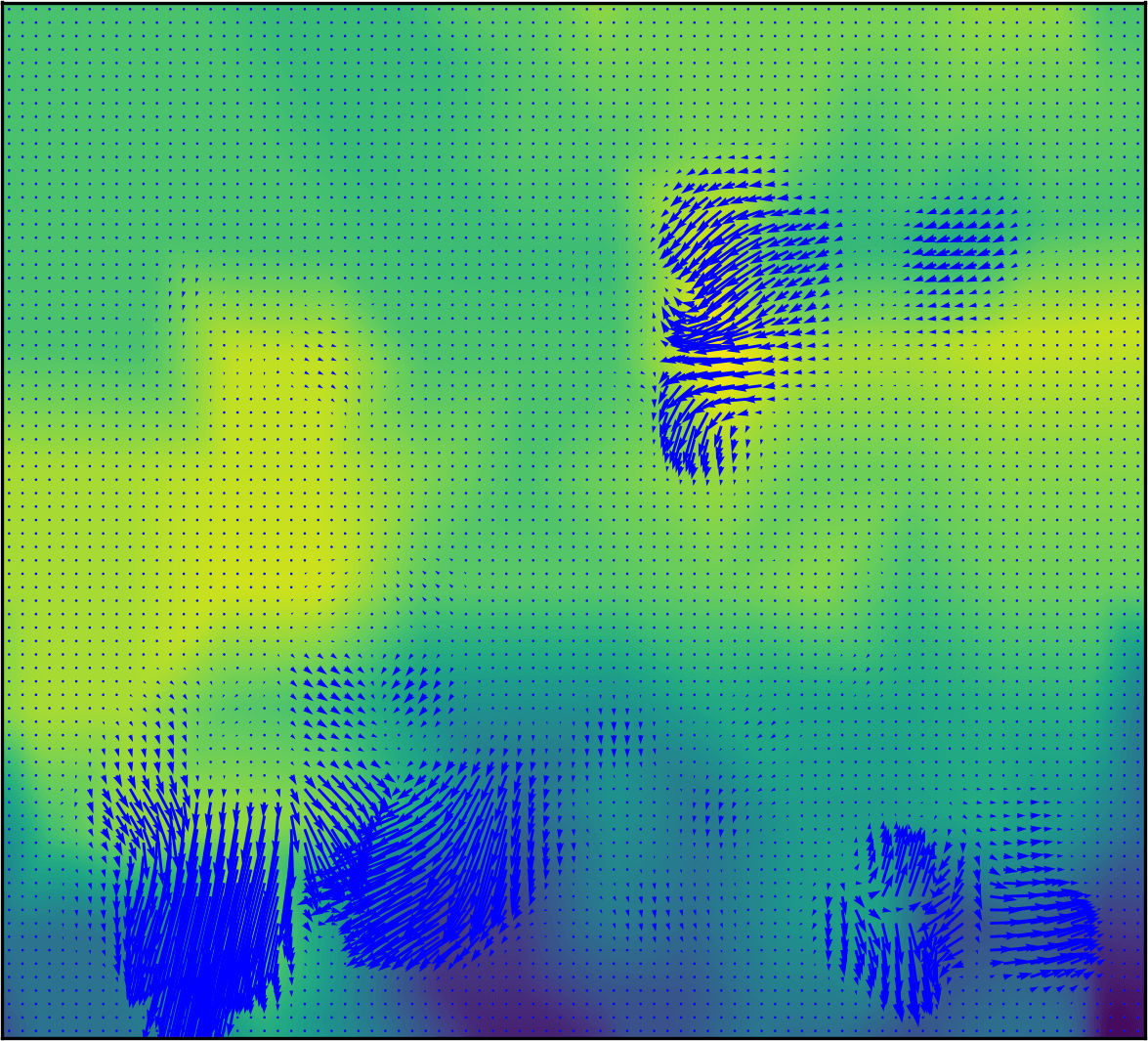}
    \end{subfigure}
    \
    \begin{subfigure}{0.22\textwidth}
        \includegraphics[width=\textwidth]{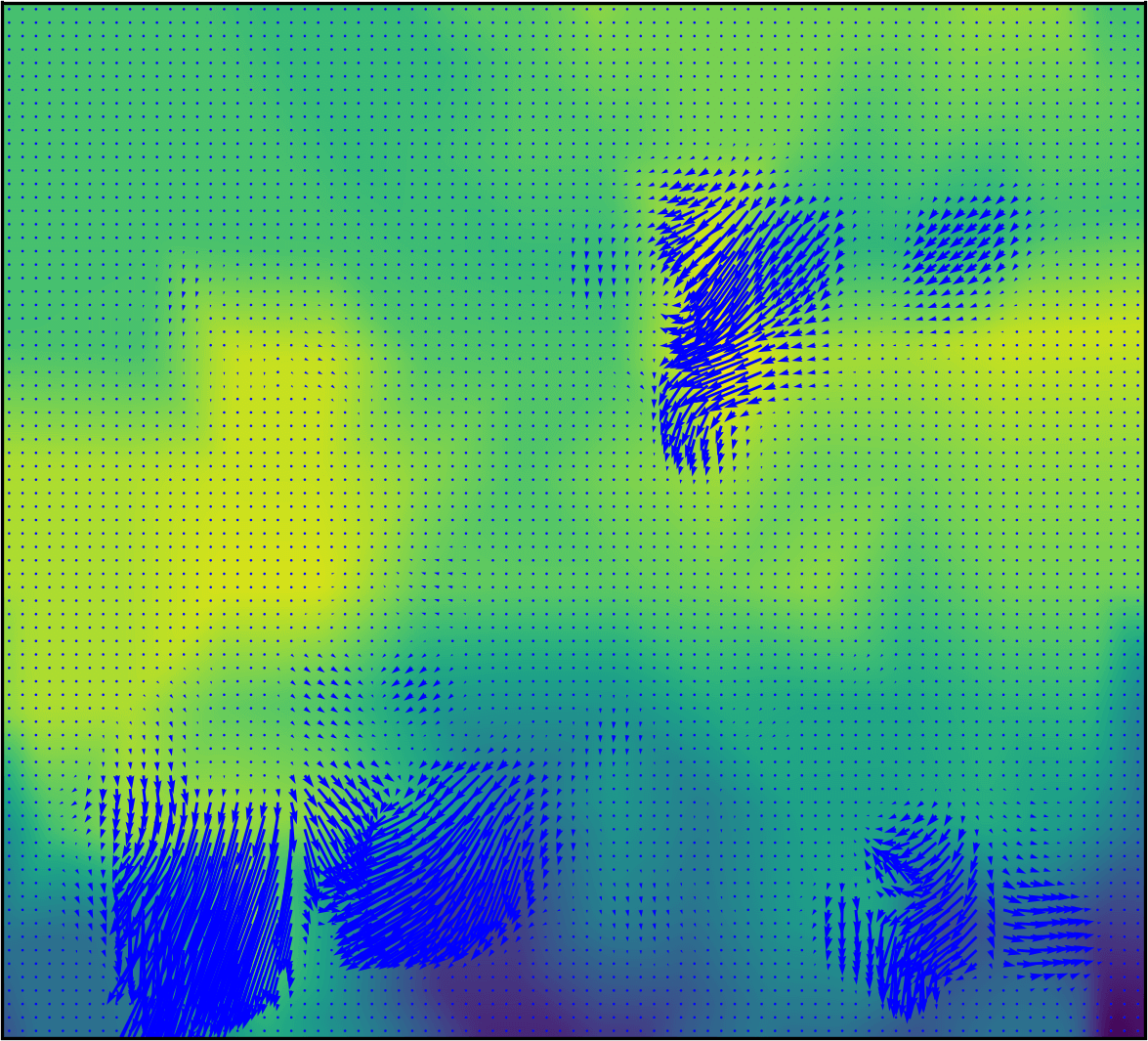}
    \end{subfigure}
    \caption{A sequence of tactile flows when the robot moves across a bump.}
    \label{fig:sample_flow_bump}
\end{figure*}
\subsection{Tactile Flow Calculation}
The impedance values over the surface of the sensor can be interpolated using~\eqref{eqn:inter3d}. These values are directly proportional to the normal pressure at taxel positions. We can draw the normal vectors of the taxels with their impedance value as their length. The endpoint of these vectors can construct a new surface as shown in Fig.~\ref{fig:3Dsurface}. This surface is analogous to a 3-dimensional surface waving in the space and we call this surface the \emph{tactile surface}.
%We claim that the tactile frame is analogous to a waving 3-dimensional surface as shown in Fig.~\ref{fig:3Dsurface}. We name this surface the \emph{tactile surface}. 

Any local changes in the impedance values on the sensor result in motion in the corresponding neighborhood of the tactile surface. Using this analogy, we define the computational tactile flow as the optical flow for the motion of the tactile surface. Here, we calculate optical flows using projections of the tactile surface on various camera planes. We name a projection of the tactile surface on a camera plane as a \emph{tactile frame}.
Depending on the camera position, we can have different tactile frames. For example, if the camera is fixed at the top of the surface and faced down toward it, the optical flow perceived by this camera is the tactile flow in $x-y$ plane. Similarly, we can find the tactile flow projected in other planes. For instance, we can project the left and right half of the tactile surface to the $y-z$ plane and find two other tactile flows in the left and right of the sensor. 

The impedance values, $p(\bm{x})$, do not remain constant and $\frac{d}{dt}p(\bm{x})\neq 0$ but the intensity of each point on the surface, $I(x,y)$, remains unchanged for spatial and time perturbations and thus for each projection of the tactile surface, the optical flow is calculated using the famous Horn and Schunck equation~\cite{horn1981determining}
\begin{equation}
    \frac{dI}{dt}=\frac{\partial I}{\partial x}v_x+\frac{\partial I}{\partial y}v_y+\frac{\partial I}{\partial t}=0,
\end{equation}
where $v=[v_x \quad v_y]^T$ is the tactile flow. In this work, we estimate the tactile flow based on Farneb\"ack polynomial expansion method~\cite{farneback2003two}. This method approximates all neighborhoods of both frames as second-order polynomials and estimates the displacements by observing the polynomial expansion coefficients~\cite{farneback2003two}. Fig.~\ref{fig:sample_frame} shows a sample tactile frame of the projection of the tactile surface on the plane beneath the sensor (on $x-y$ plane). Fig.~\ref{fig:sample_flow_proj} shows the tactile flow for this frame. Here, the Shadowhand is moving on a flat surface. %Fig.~\ref{fig:sample_sequence} shows the tactile flows for to the motion of the robot on the surface which has a small round bump on it.
%Here we see a radial tactile flow which relates to pushing the tactile flow towards the surface.
%%%%%%%%%%%%%%%%%%%%%%%%%%%%%%%%%%%%%%%%%%%%%%%%%%%%%%%%%%%%%%%%%%%%%%%%%%%%%%%%
\section{Experiments}\label{sec:experiments}
\subsection{Experimental Setup}
\begin{figure}[h!]
    \centering
    \includegraphics[scale=1.0]{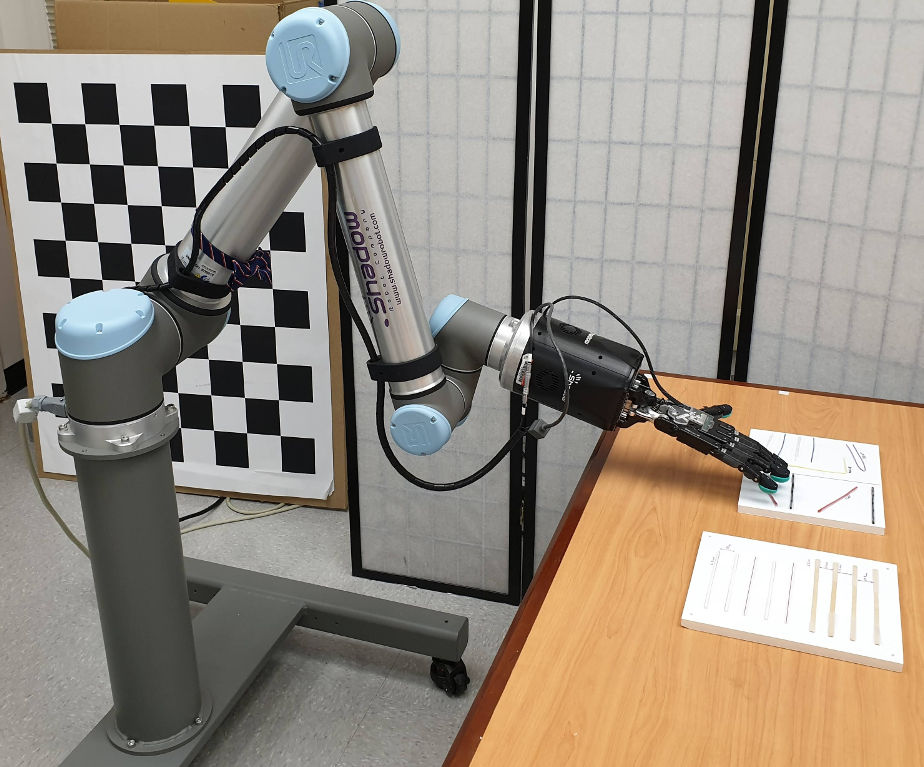}
    \caption{Our experimental setup showing the manipulator with the end-effector}
    \label{fig:setup}
\end{figure}

As mentioned earlier in Sec.~\ref{sec:introduction}, tactile flow can be classified under two main classes, namely \textit{static} and \textit{dynamic}. Static tactile flow encodes flow patterns in situations when the sensor and the object remain almost stationary in contact, but the skin deforms due to static friction between the surfaces in contact. Dynamic tactile flow encodes information in scenarios where the finger has already overcome the static frictional forces and is moving over a surface. In both cases, the flow patterns encode information about the direction of motion, the relative magnitude of forces being applied as well as latent shape information.
To demonstrate our approach, we perform several experiments on our custom-designed surfaces in order to best validate our hypotheses on using tactile flow and overall pressure to discover information about the surface.

Our hardware setup, shown in Fig.~\ref{fig:setup} includes a UR10 robotic manipulator, on which is mounted the Shadow Dexterous Hand as the end-effector. There is a table placed in front of the manipulator and end-effector setup, on which are mounted the various experiment boards. The robot is controlled through Robotic Operating System (ROS). The robot's position, as well as the end-effector's fingers, are tracked using the \textit{tf}\cite{rostf} library provided in ROS.

\begin{figure}[h!]
    \centering
    \begin{subfigure}{0.225\textwidth}
        \includegraphics[width=\textwidth]{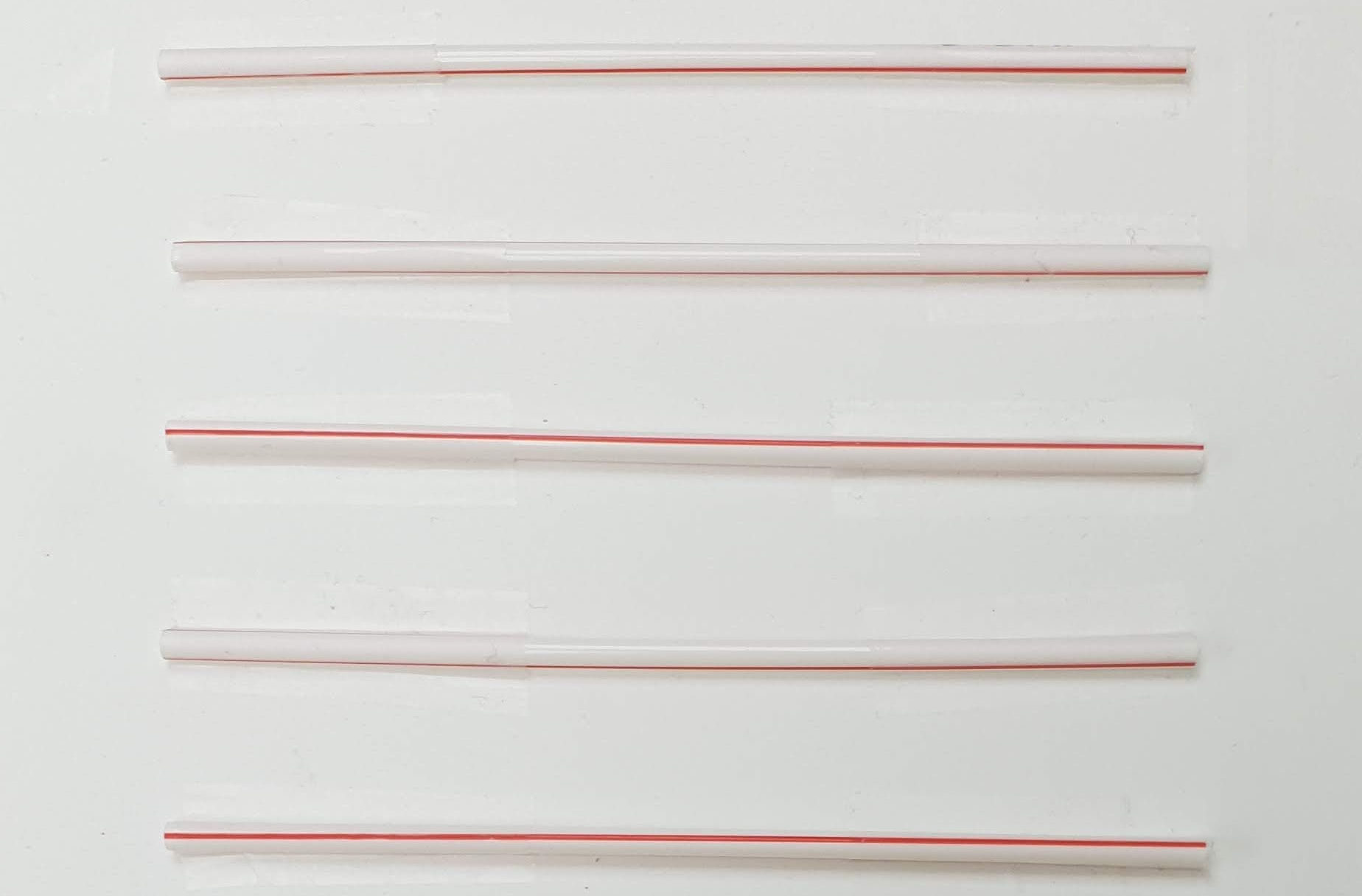}
        \caption{Straws}
        \label{fig:exp_straws}
    \end{subfigure}
    \quad
    \begin{subfigure}{0.225\textwidth}
        \includegraphics[width=\textwidth]{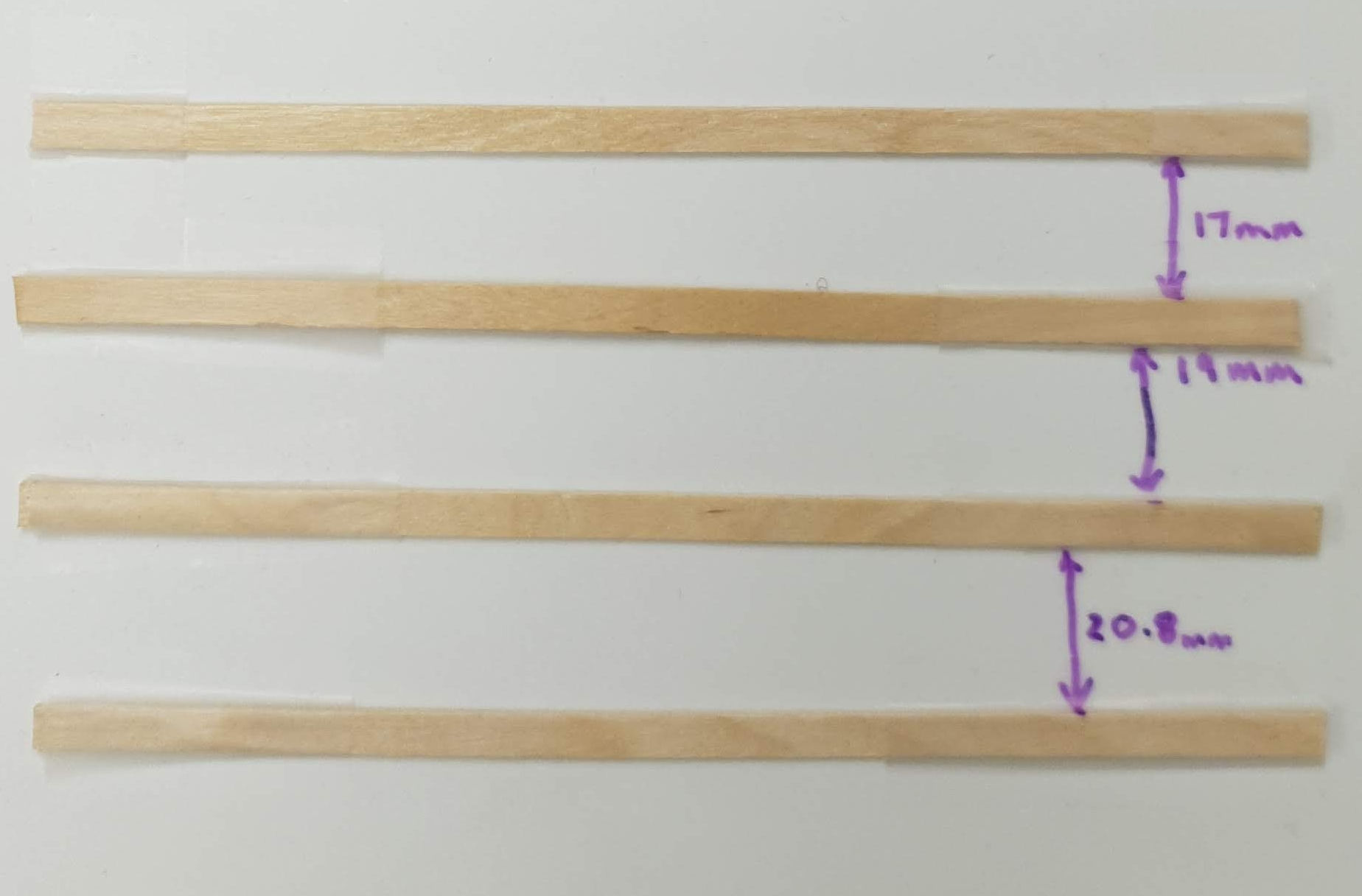}
        \caption{Sticks}
        \label{fig:exp_sticks}
    \end{subfigure}
\caption{Experimental textured surfaces}
\label{fig:exps_surface}
\end{figure}

For our experimental surfaces, we designed several boards on which we mounted different textured objects such as wires of varying thickness and wooden sticks. Examples of our experimental surfaces can be seen in Fig.~\ref{fig:exps_surface}. 

\subsection{Dynamic Tactile Flow}
\begin{figure}[h!]
    \centering
    \begin{subfigure}{0.225\textwidth}
        \includegraphics[width=\textwidth]{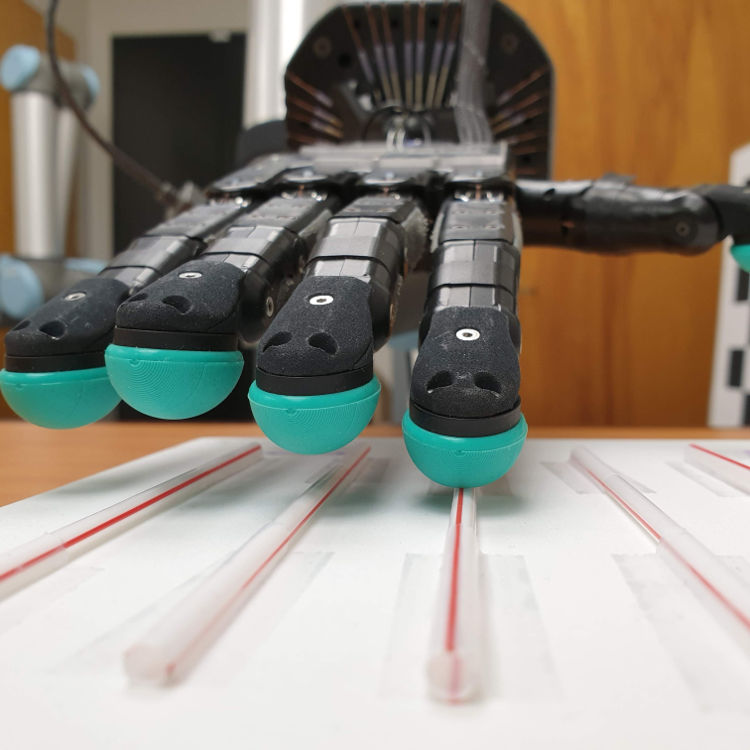}
        \caption{Finger moving over the straw}
        \label{fig:exp_straws_side}
    \end{subfigure}
    \quad
    \begin{subfigure}{0.225\textwidth}
        \includegraphics[width=\textwidth]{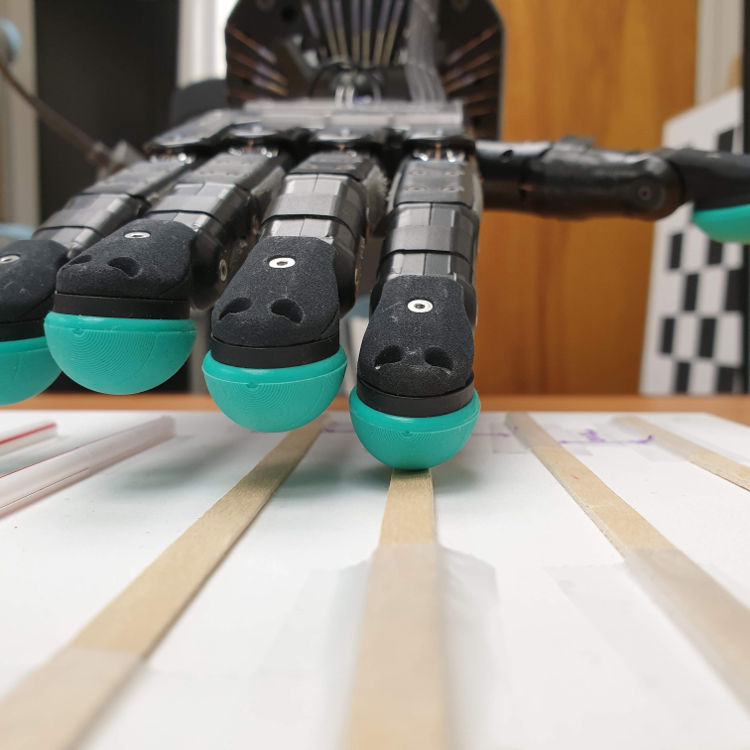}
        \caption{Finger moving over the stick}
        \label{fig:exp_sticks_side}
    \end{subfigure}
\caption{Finger moving over different textured surfaces}
\label{fig:exps_surface_side}
\end{figure}

The surfaces shown in Figs.~\ref{fig:exp_straws} and \ref{fig:exp_sticks} respectively have been carefully chosen to reflect different circumstances a finger might encounter. The straws surface is used to detect high and low surface differences and distinguish between angles of approach and departure. This is facilitated by the fact that at the time of contact, the finger ``lifts off'' the flat surface and only the part of the finger in contact with the straw surface is ``activated''. By comparing the computed flow, we are able to detect the direction from which the finger moved over the straw as well as measure the relative height of the straw by looking at the pressure peaks. This is demonstrated in Fig.~\ref{fig:exp_straws_side}.

A similar experiment is performed with the ``sticks surface'', which is designed to have a lower height difference to the ground surface but is wider than the straws. This is done to compare and contrast the readings in pressure and tactile flow between the two and demonstrate the utility of tactile flow in detecting the direction of motion over different surfaces. In this case, we are also able to estimate the width of the stick by analyzing the pressure plot.

\begin{figure}
    \centering
    \includegraphics[width=0.5\textwidth]{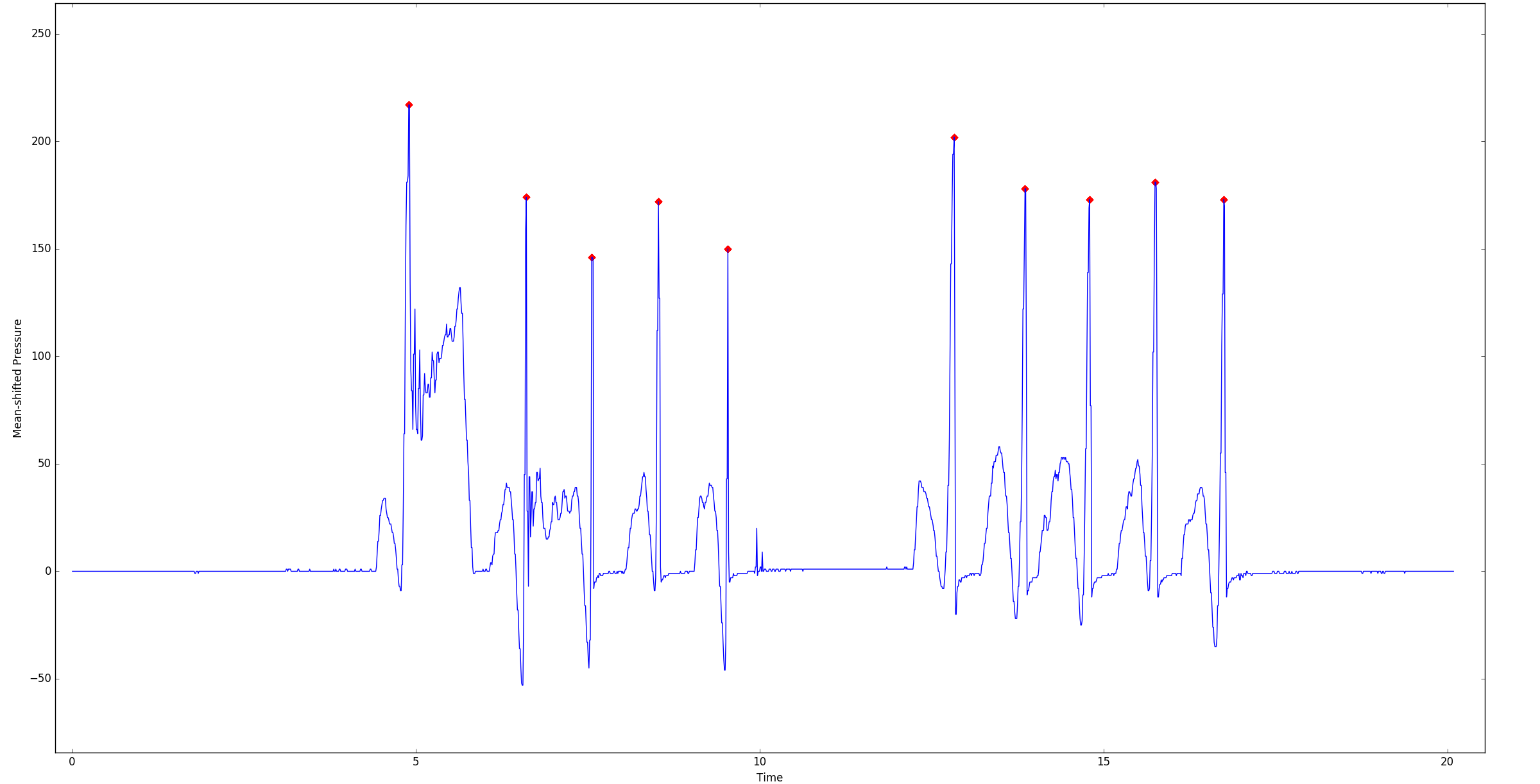}
    \caption{Peaks in Mean-Shifted Pressure}
    \label{fig:pressure_peaks}
\end{figure}

The BioTac sensors provide both pressure and impedance readings, and we utilize both in our experiments. While the impedance values are used to construct the ``tactile surface'' described above, the pressure readings are useful to find regions of interest during the finger's movement over textured surfaces. We can use the peaks from the pressure readings in order to identify where significant events, such as bumps, approaching or departing an edge, etc. occurs and use those as priors for selecting frames for flow computation. This is facilitated by the fact that all our data is time-synchronized using a common clock.
One such example is shown in Fig.~\ref{fig:pressure_peaks} for the \textit{straws} sequence.

\subsubsection{Sticks Sequence}
\begin{figure}[h!]
    \centering
    \begin{subfigure}{0.225\textwidth}
        \includegraphics[width=\textwidth]{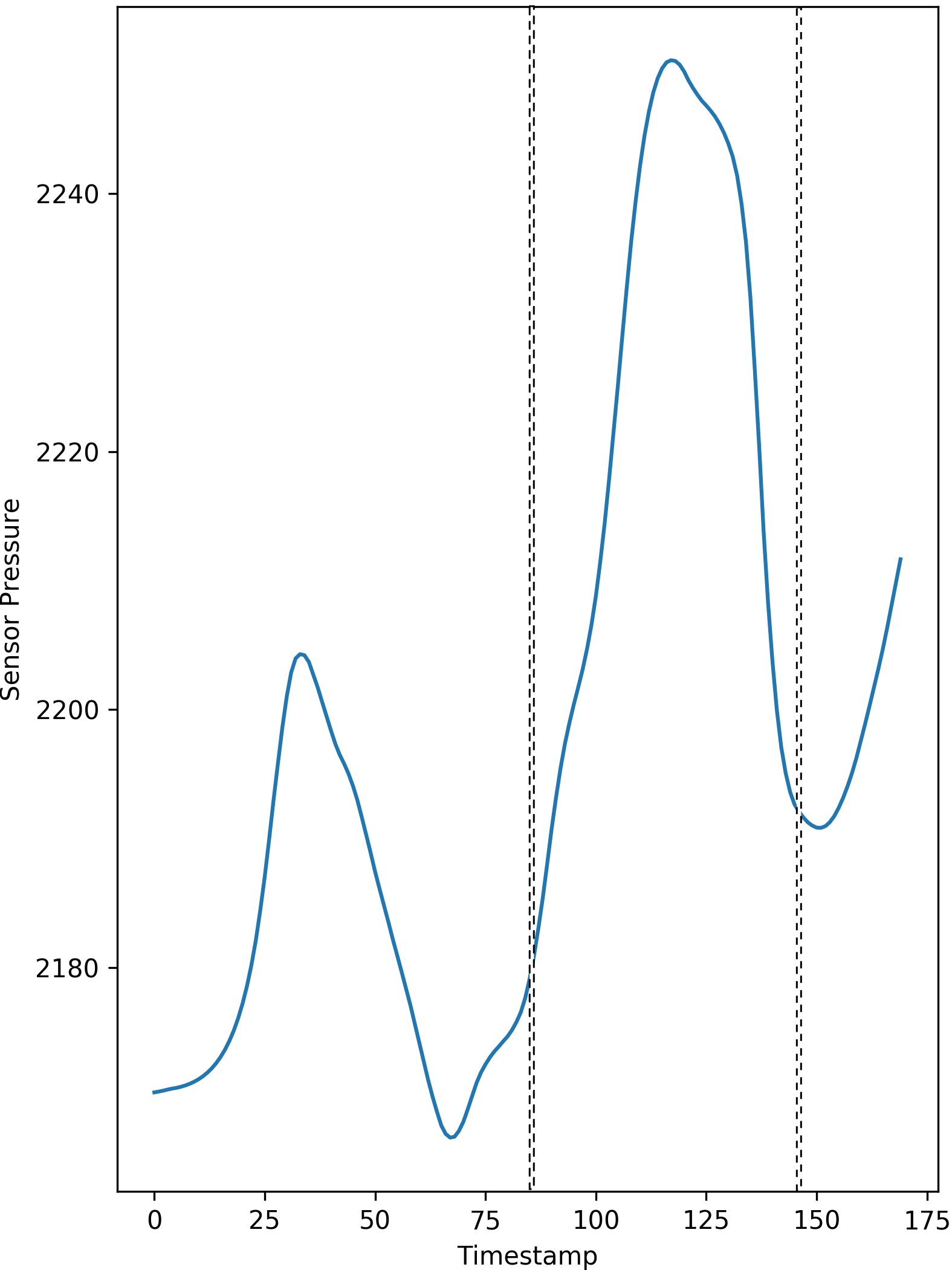}
        \caption{Sensor Pressure}
        \label{fig:sticks_pressure}
    \end{subfigure}
    \quad
    \begin{subfigure}{0.225\textwidth}
        \includegraphics[width=\textwidth]{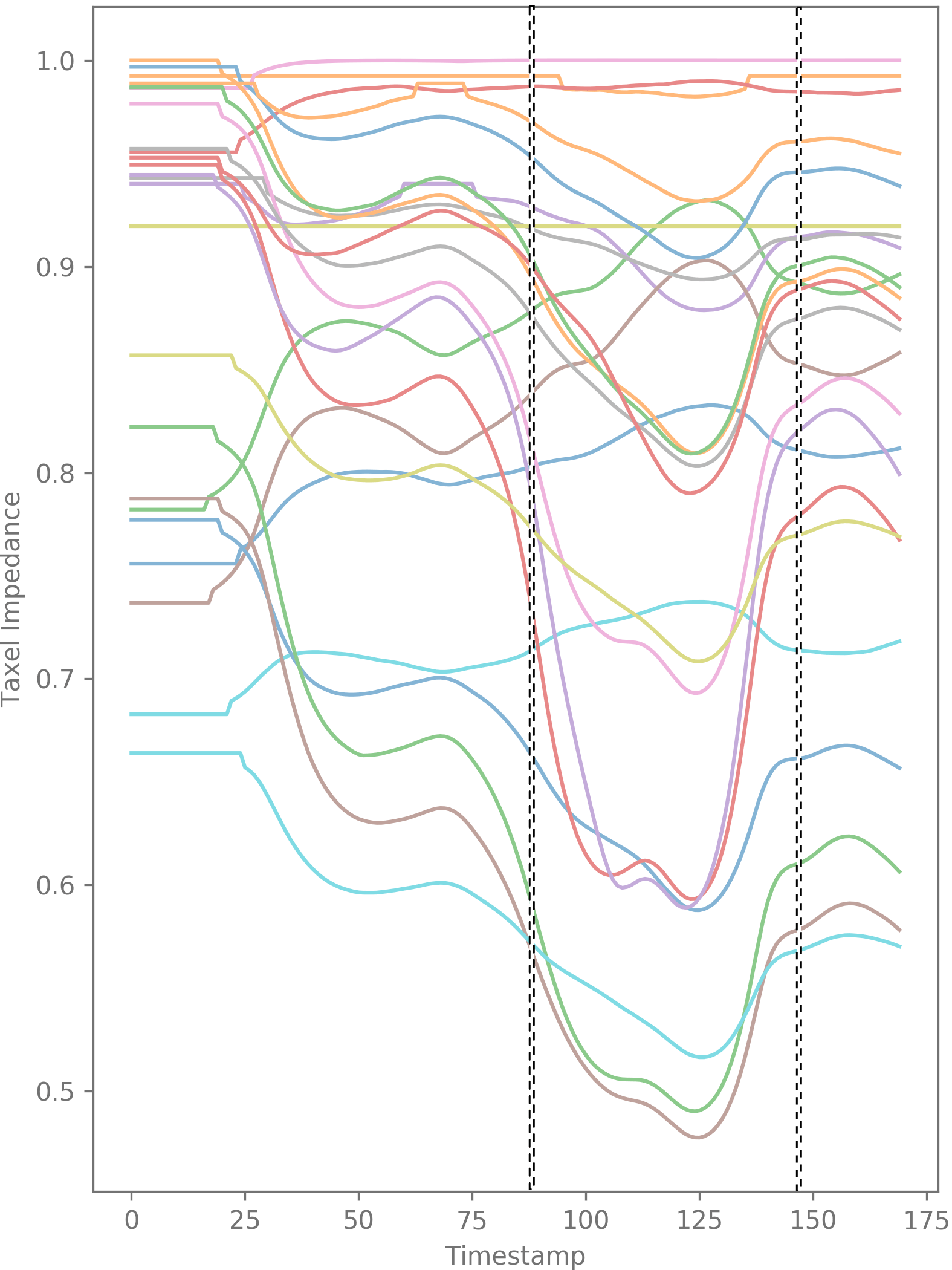}
        \caption{Taxel Impedances}
        \label{fig:sticks_impedance}
    \end{subfigure}
\caption{The pressure and impedance plots for the finger moving over sticks}
\label{fig:stick_plots}
\end{figure}

\begin{figure}[h!]
    \centering
    \begin{subfigure}{0.225\textwidth}
        \includegraphics[width=\textwidth]{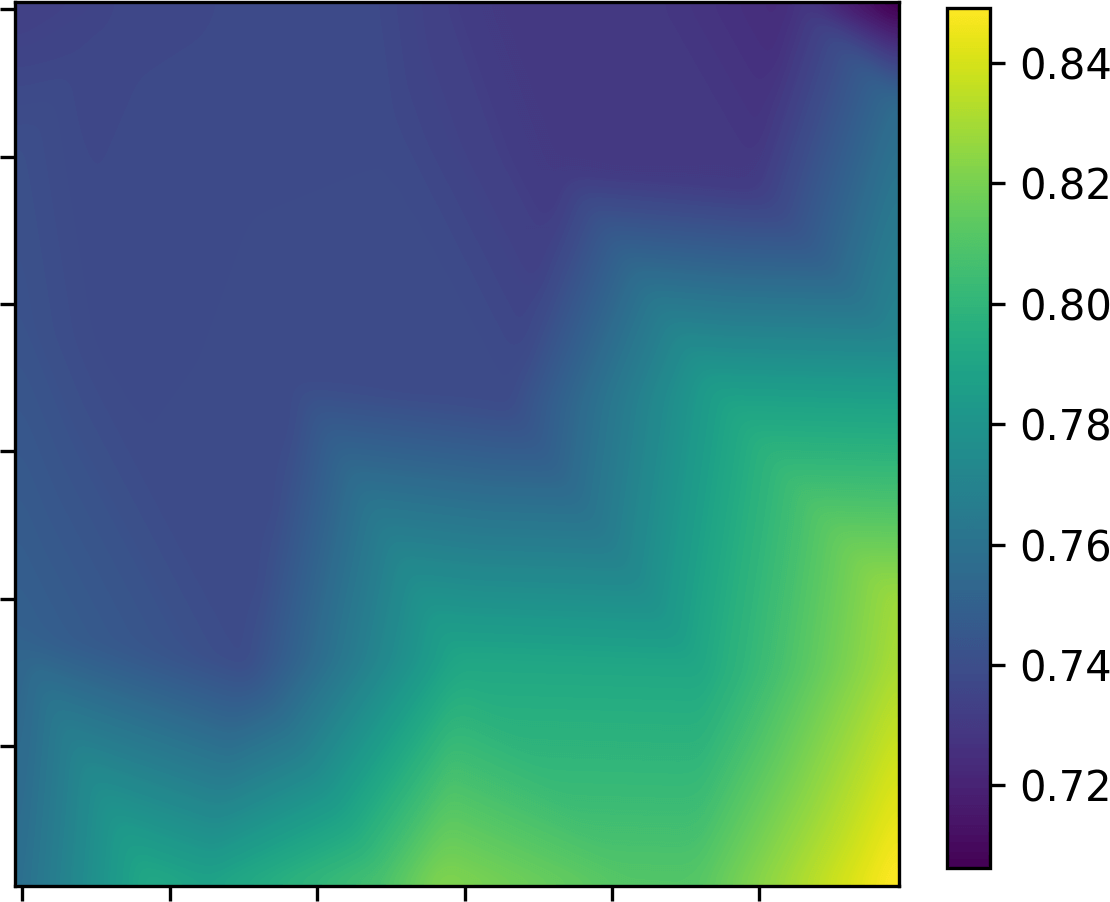}
        \label{fig:projection2_01}
    \end{subfigure}
    \quad
    \begin{subfigure}{0.225\textwidth}
        \includegraphics[width=\textwidth]{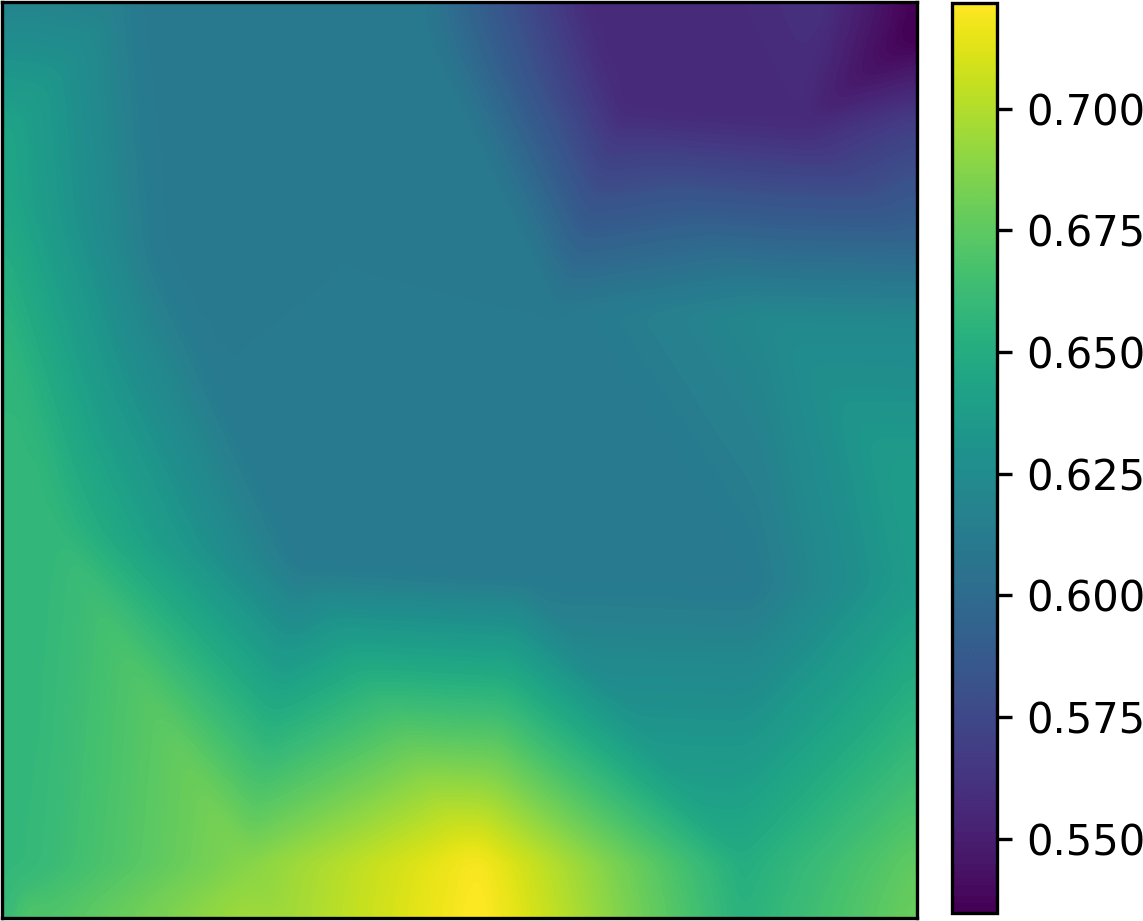}
        \label{fig:projection2_02}
    \end{subfigure}
\vspace{-17px}
\caption{2D projections of 3D taxel impedances for flow computation}
\label{fig:stick_projections}
\end{figure}

\begin{figure}[h!]
\centering
    \begin{subfigure}{0.5\textwidth}
        \centering
        \includegraphics[width=0.45\textwidth]{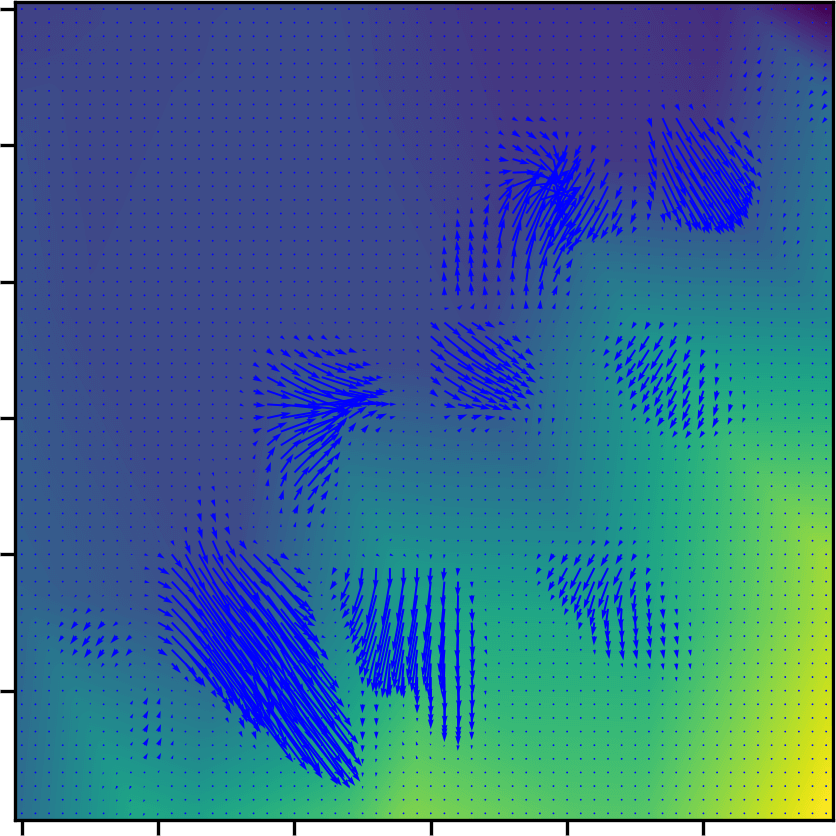}%
        \quad
        \includegraphics[width=0.45\textwidth]{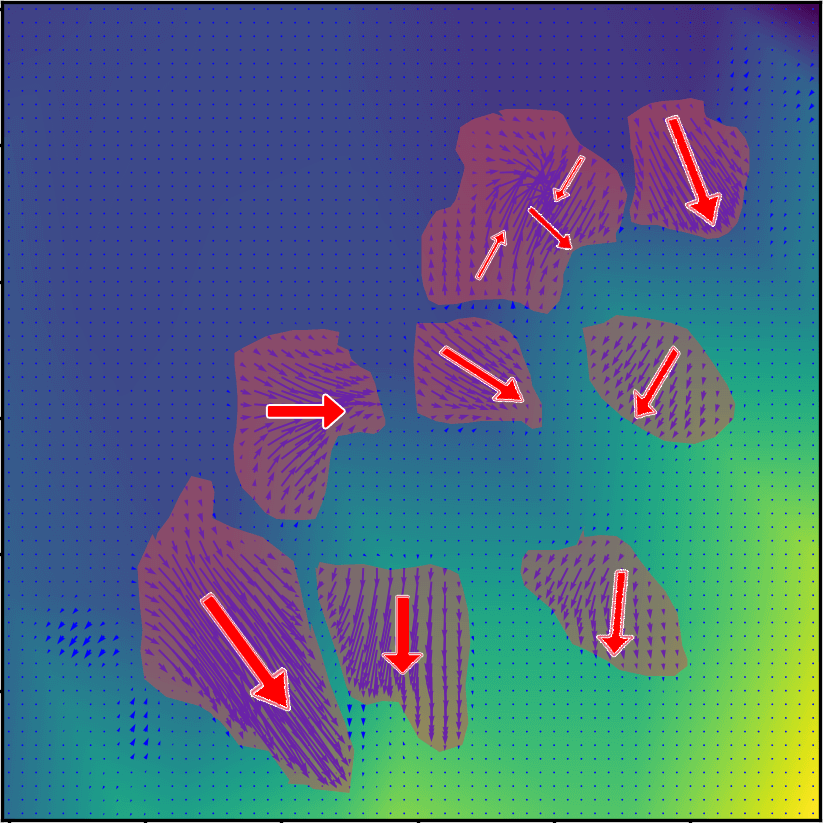}
        \caption{Before}
    \end{subfigure}
    \\
    \begin{subfigure}{0.5\textwidth}
        \centering
        \includegraphics[width=0.45\textwidth]{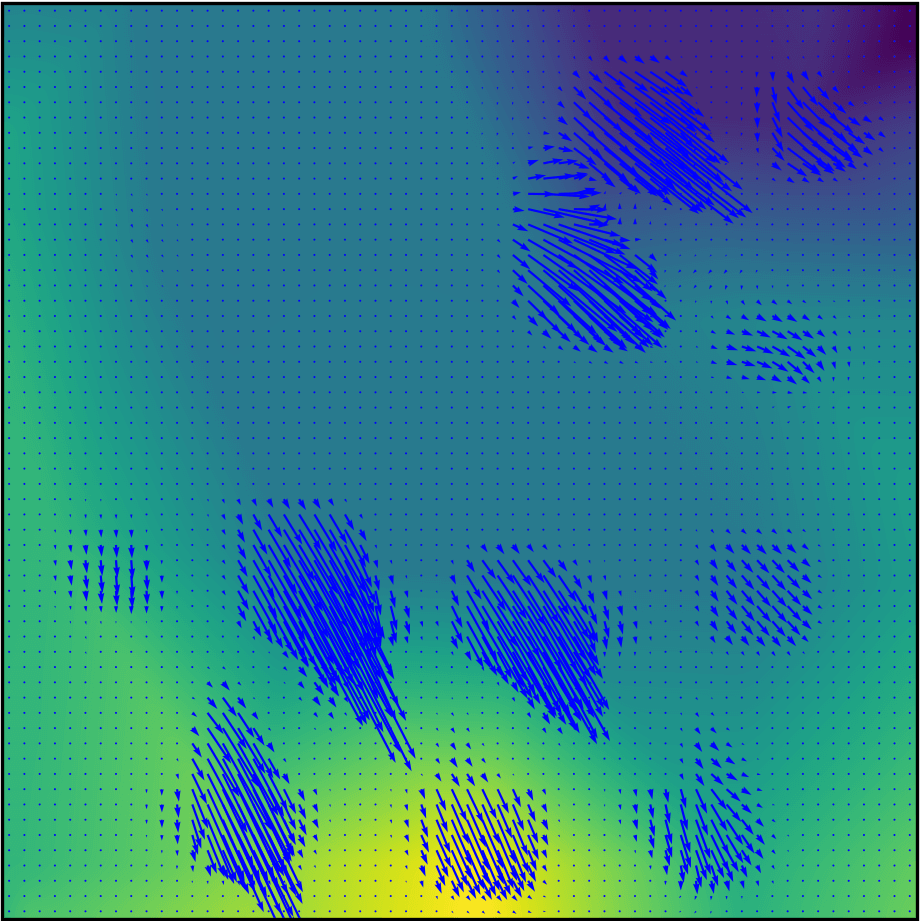}%
        \quad
        \includegraphics[width=0.45\textwidth]{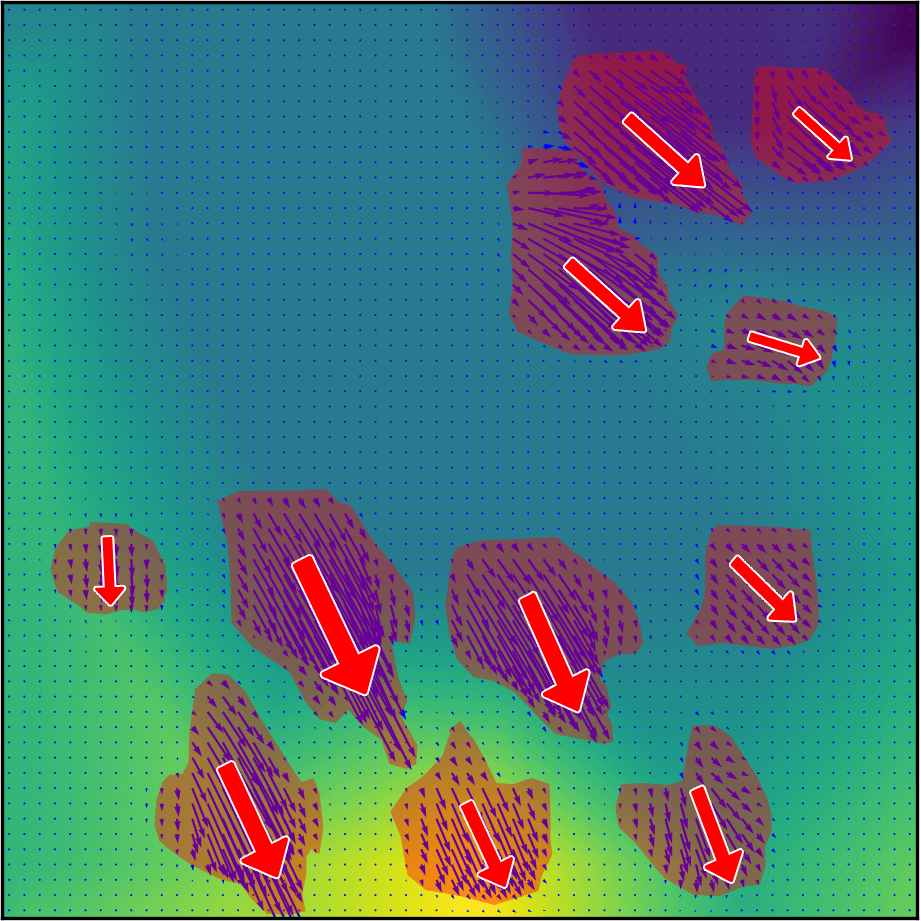}
        \caption{During}
    \end{subfigure}
    \\
    \begin{subfigure}{0.5\textwidth}
        \centering
        \includegraphics[width=0.45\textwidth]{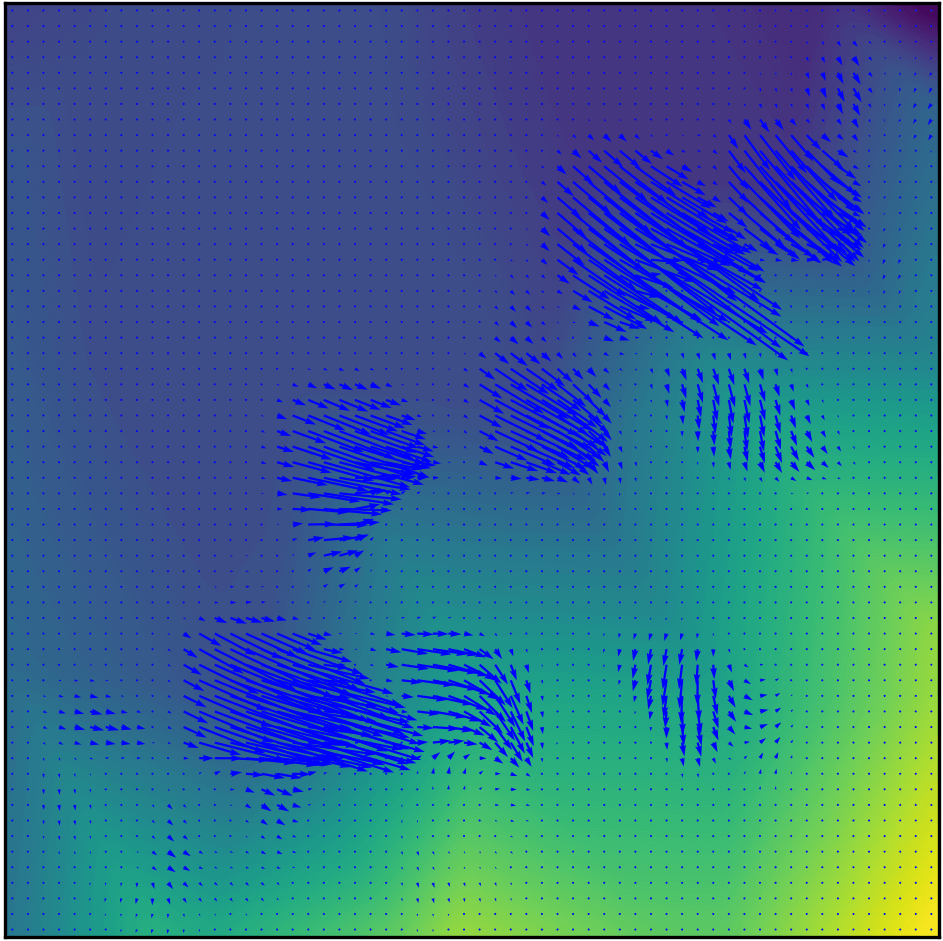}%
        \quad
        \includegraphics[width=0.45\textwidth]{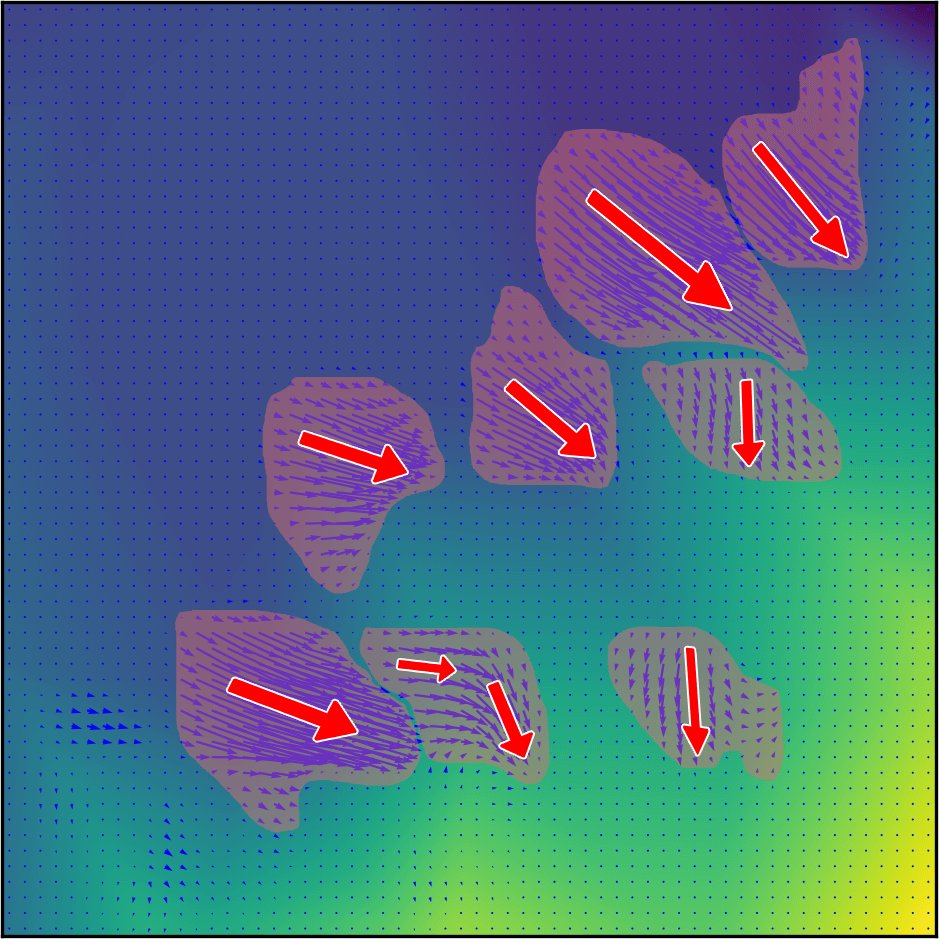}
        \caption{After}
    \end{subfigure}
    \caption{3 tactile flow images and corresponding aggregated flow for the Sticks sequence}
    \label{fig:sticks_flow_seq}
\end{figure}

As can be seen in Fig.~\ref{fig:stick_plots}, the bands point us towards the region of interest (sequence of frames) which has been obtained from the pressure plot (Fig.~\ref{fig:sticks_pressure}). The corresponding bands in the impedance plot show similar peaks and troughs, and we can thus focus on computing tactile flow in those specific regions only. As described in our method, Fig.~\ref{fig:stick_projections} shows two sample projections from the 3D ellipsoidal model to a 2D image. For our flow computations, we pick three distinct regions, namely one just before contact as the finger approaches the stick, one region as the finger is moving over the stick and one just after the finger has left the stick and is moving away. The computed flow in these respective regions are representative of different ``classes'' of motion and facilitate texture and motion classification that may be performed from the tactile flow data. This is elucidated in Fig.\ref{fig:sticks_flow_seq}. Alongside, we also show aggregated flow directions, computed from the generated flow fields. This is done to better visualize the direction of motion and easier parsing of the tactile flow data.

\subsubsection{Straws Sequence}
\begin{figure}[h!]
    \centering
    \begin{subfigure}{0.225\textwidth}
        \includegraphics[width=\textwidth]{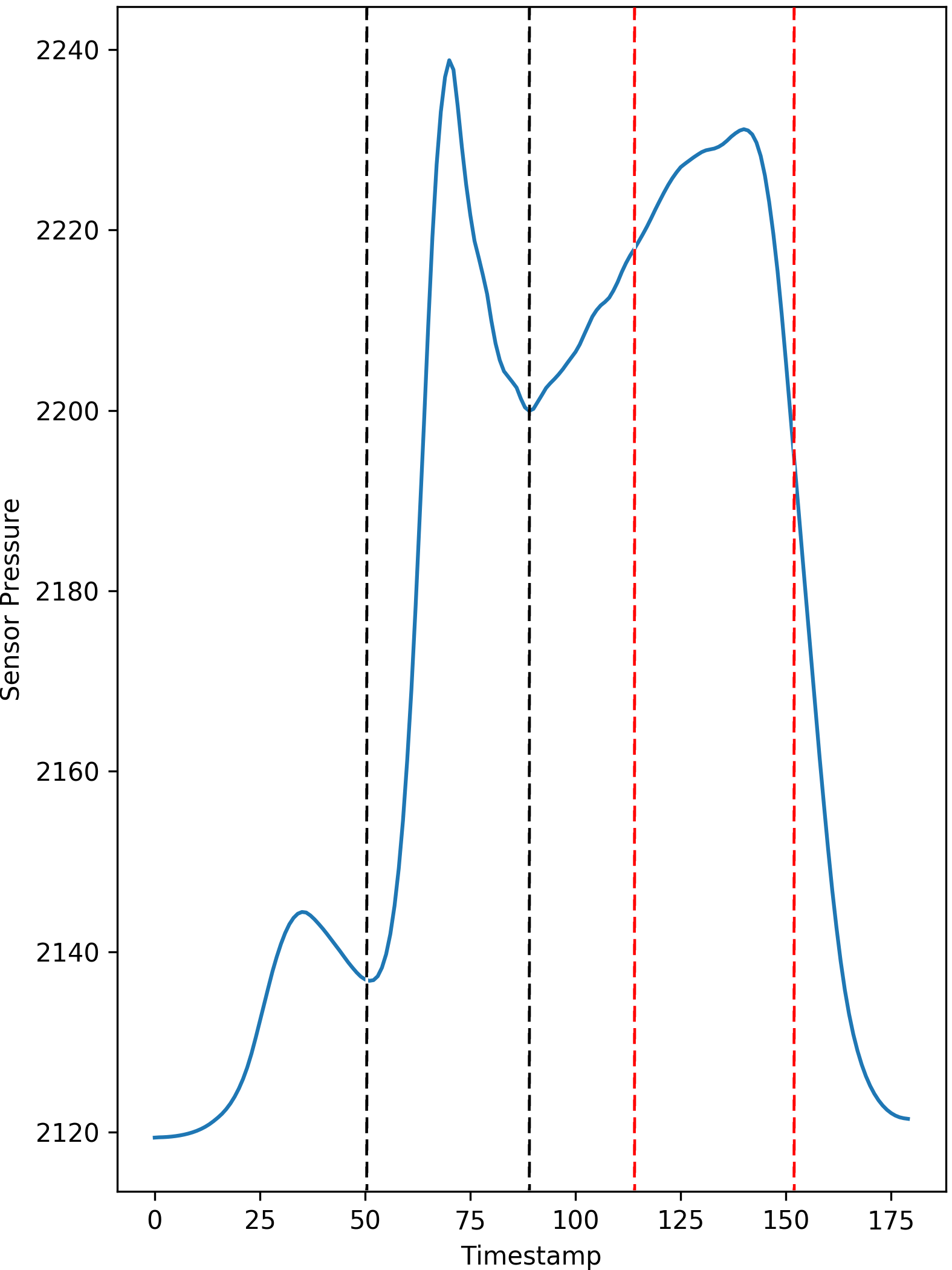}
        \caption{Sensor Pressure}
        \label{fig:straws_pressure}
    \end{subfigure}
    \quad
    \begin{subfigure}{0.225\textwidth}
        \includegraphics[width=\textwidth]{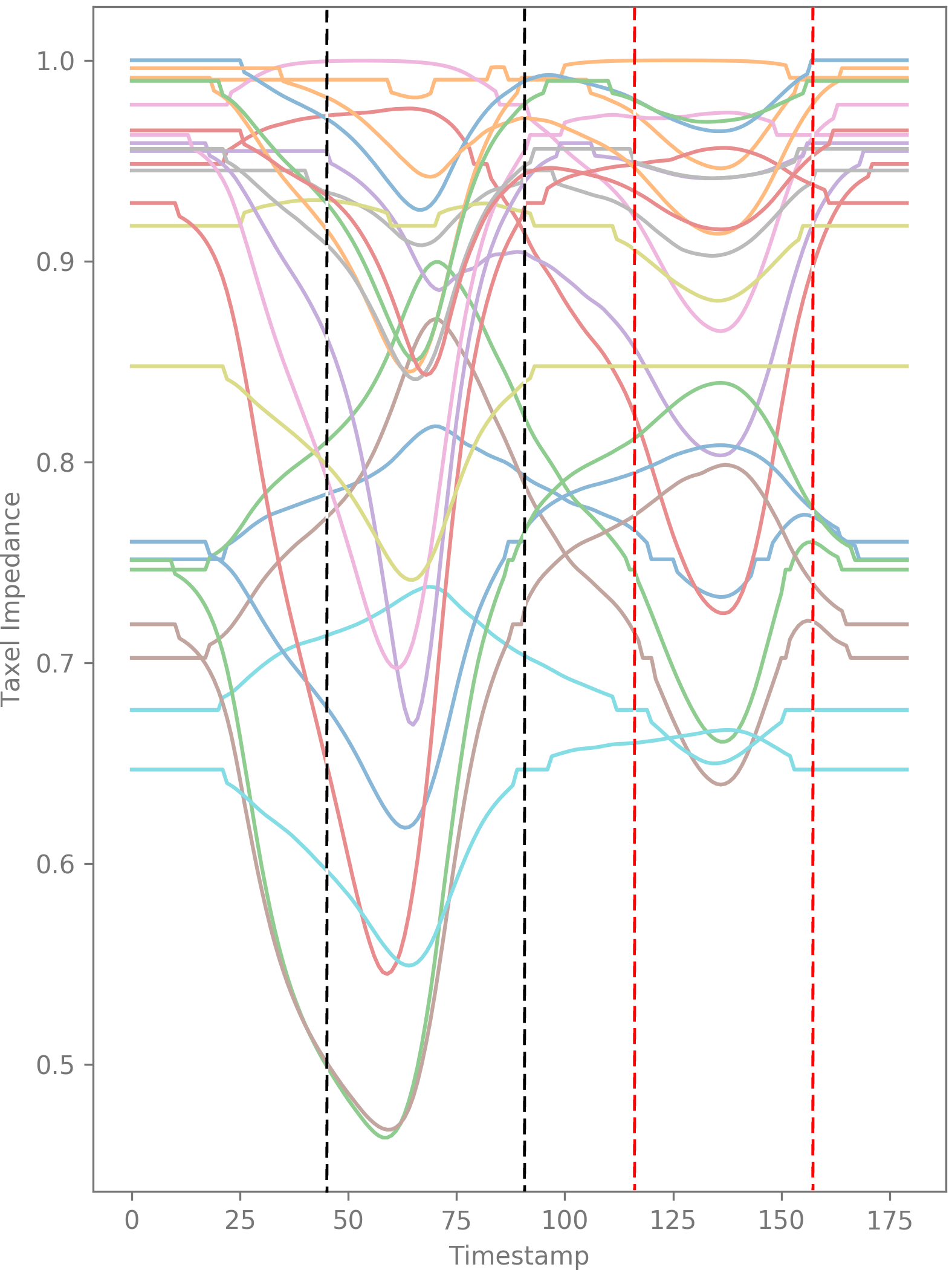}
        \caption{Taxel Impedances}
        \label{fig:straws_impedance}
    \end{subfigure}
\caption{The pressure and impedance plots for the finger moving over straws}
\label{fig:straw_plots}
\end{figure}

\begin{figure}[h!]
\centering
     \begin{subfigure}{0.5\textwidth}
        \centering
        \includegraphics[width=0.45\textwidth]{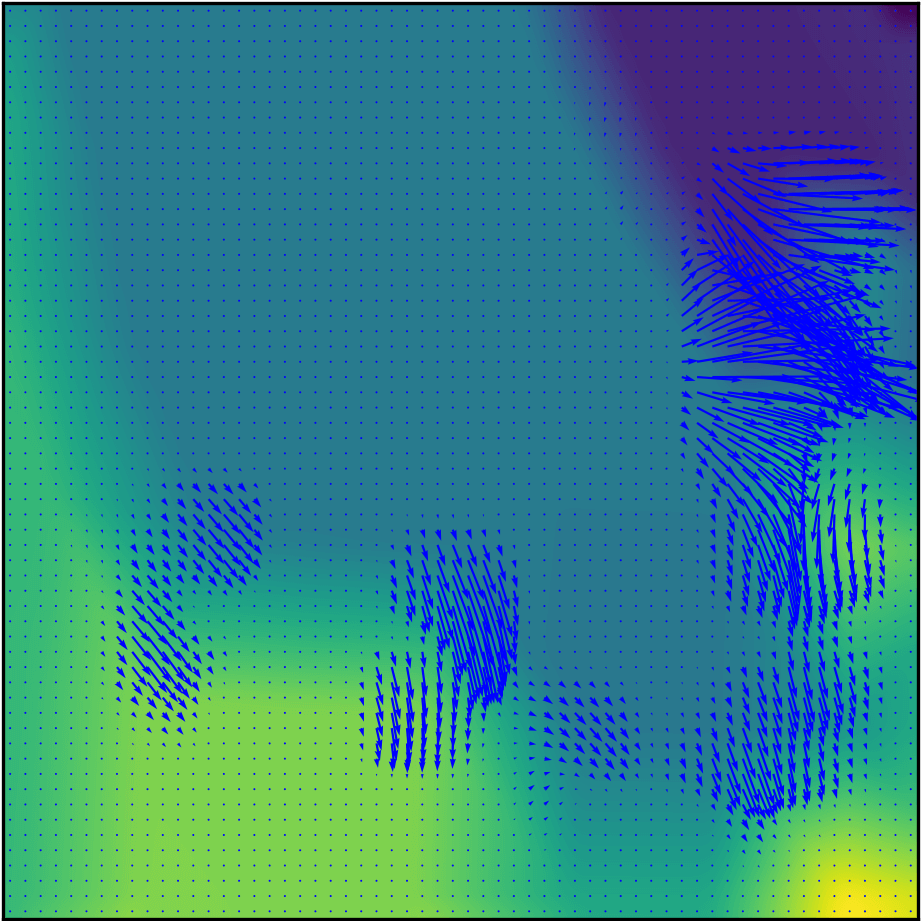}%
        \quad
        \includegraphics[width=0.45\textwidth]{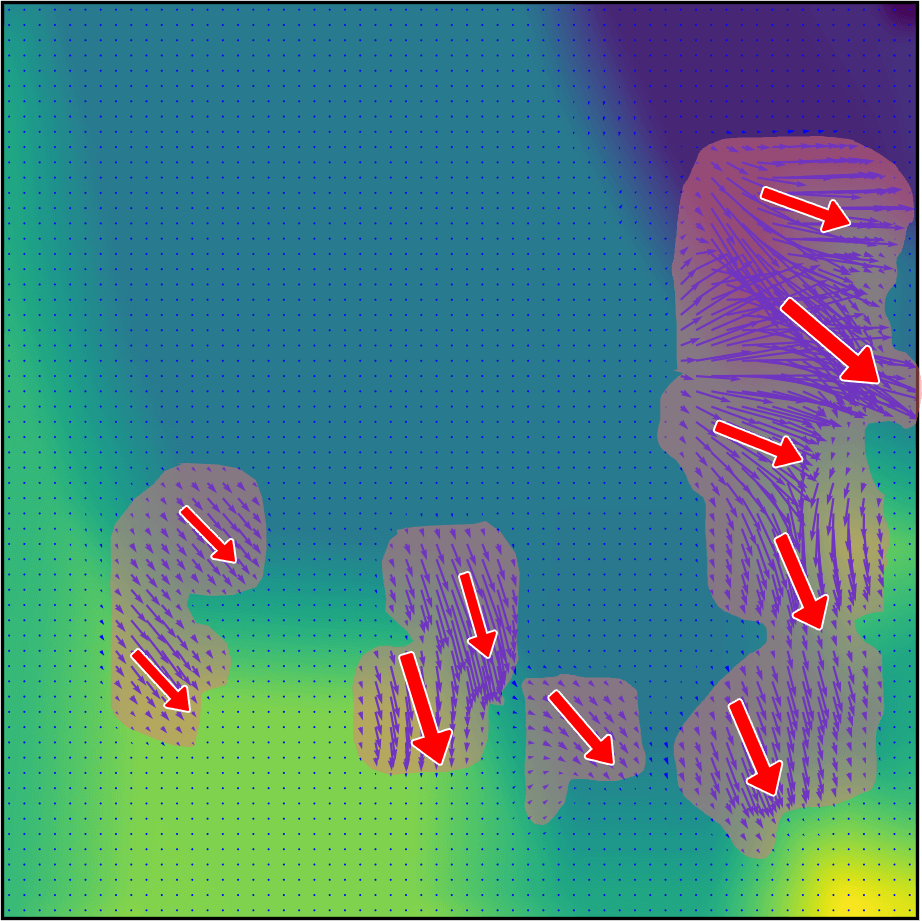}
        \caption{Before}
    \end{subfigure}
    \\
    \begin{subfigure}{0.5\textwidth}
        \centering
        \includegraphics[width=0.45\textwidth]{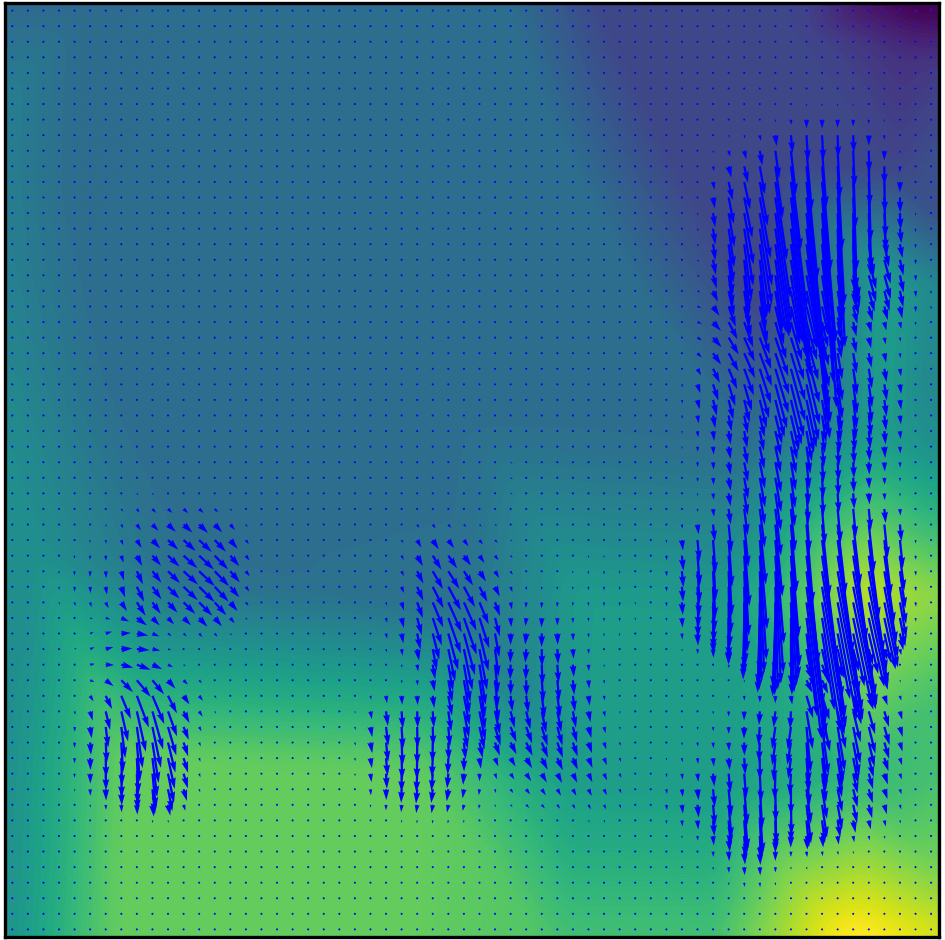}%
        \quad
        \includegraphics[width=0.45\textwidth]{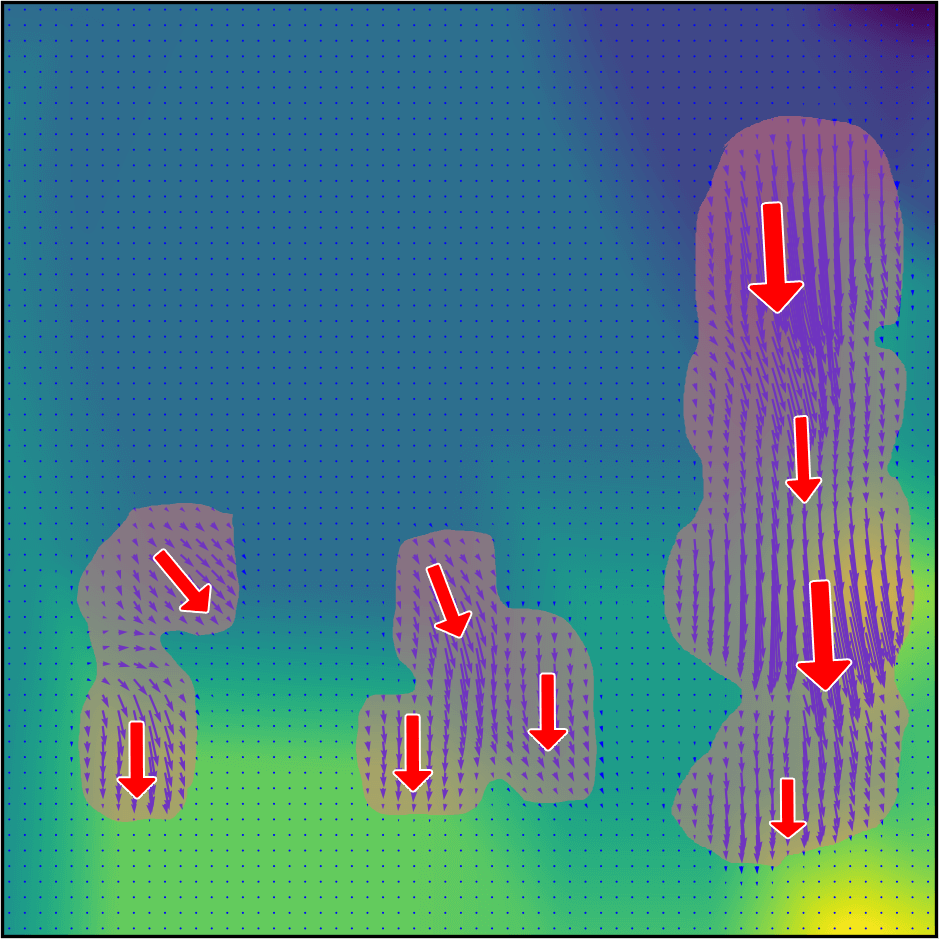}
        \caption{During}
    \end{subfigure}
    \\
    \begin{subfigure}{0.5\textwidth}
        \centering
        \includegraphics[width=0.45\textwidth]{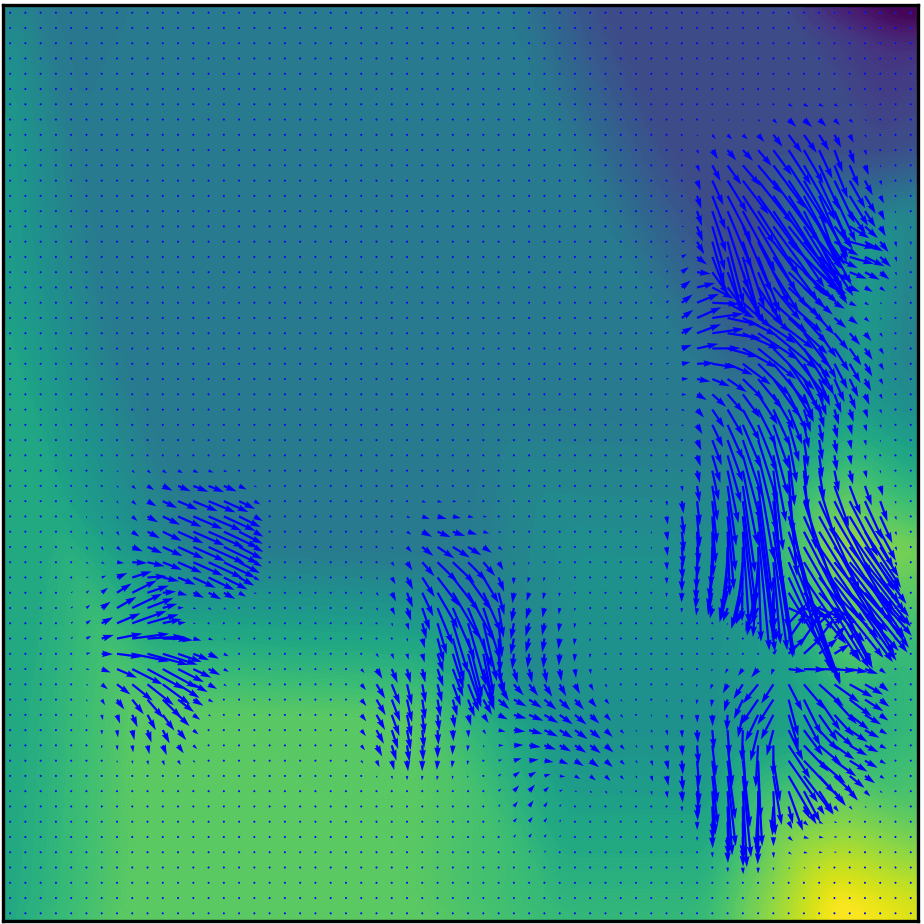}%
        \quad
        \includegraphics[width=0.45\textwidth]{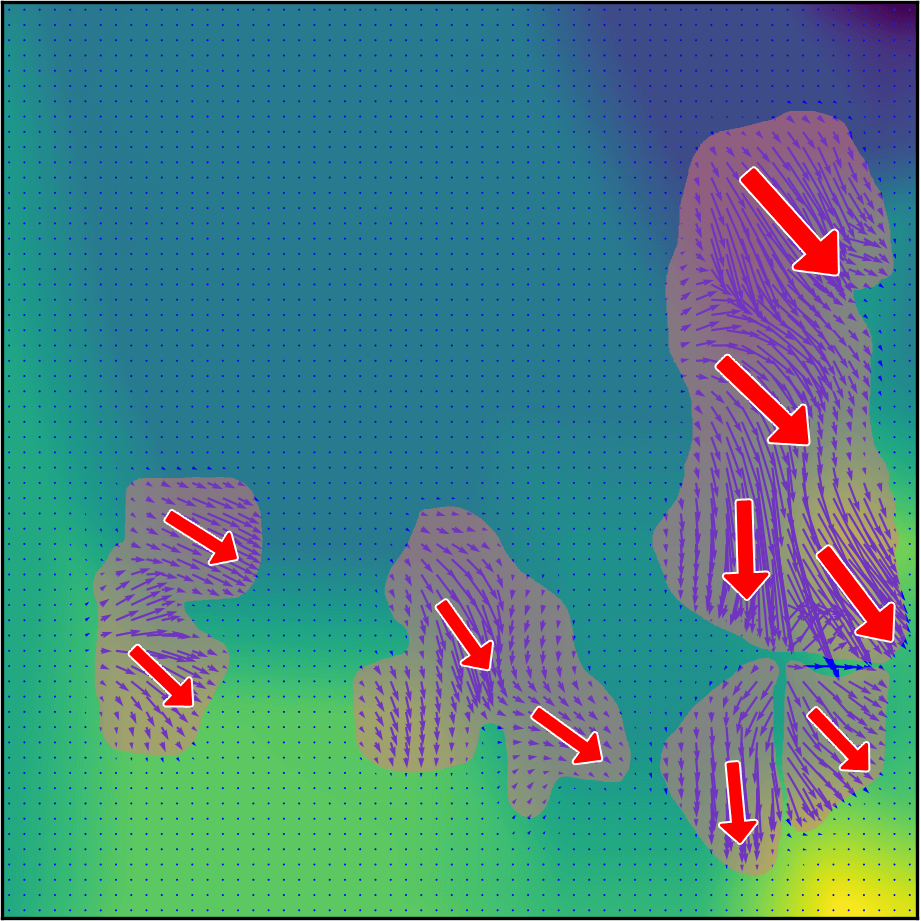}
        \caption{After}
    \end{subfigure}
    \caption{3 tactile flow images and corresponding aggregated flow for the Straws sequence}
    \label{fig:straws_flow_seq}
\end{figure}

Similar to the \textit{Sticks} sequence, for the \textit{Straws} sequence, the time-synchronized taxel impedances and the pressure values are shown in Fig.\ref{fig:straw_plots}. As before, we use the pressure peaks to find regions of interest for which we compute the tactile flow sequences. This is illustrated in Fig.~\ref{fig:straws_flow_seq}. These sequences also correspond to before, during and after the finger moves on the straw. The aggregated flow field directions are also provided.

\subsection{Static Tactile Flow}
\begin{figure}[h!]
    \centering
    \includegraphics[width=0.45\textwidth]{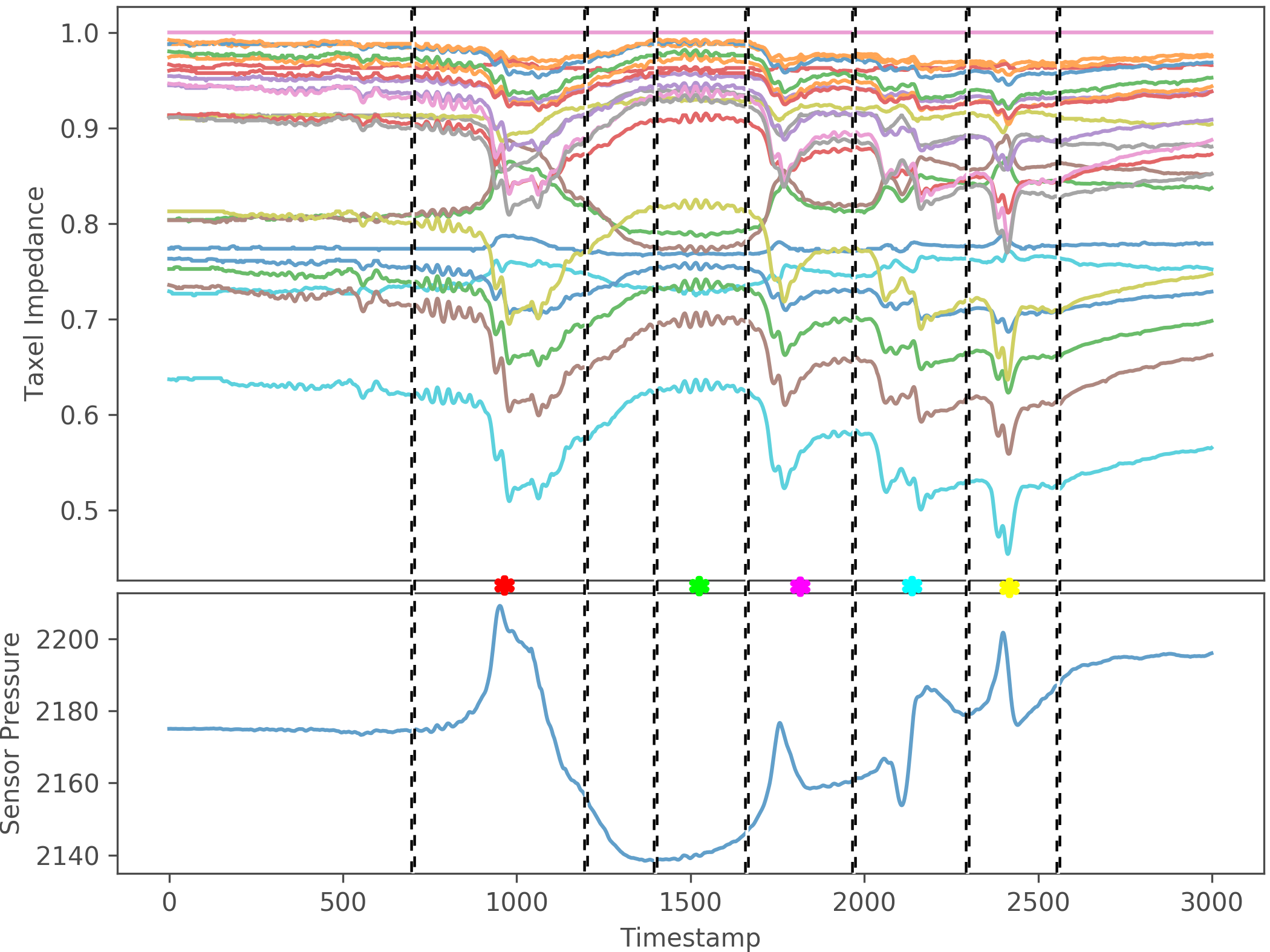}
    \caption{Combined taxel impedance and pressure plot}
    \label{fig:pour_pressures}
\end{figure}
\begin{figure*}[h!]
\centering
    \begin{subfigure}{0.5\textwidth}
        \centering
        \includegraphics[width=0.45\textwidth]{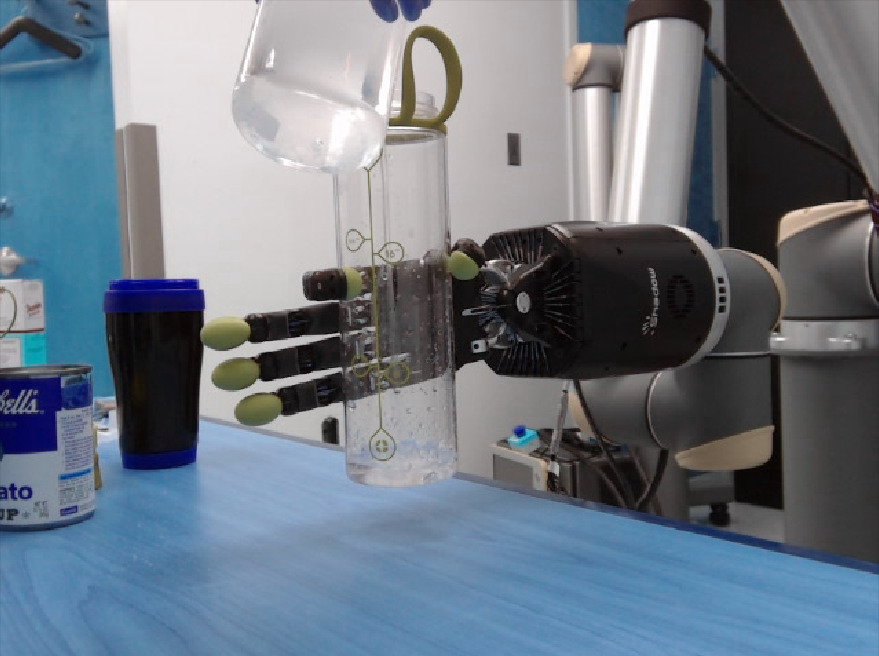}%
        \quad
        \includegraphics[width=0.45\textwidth]{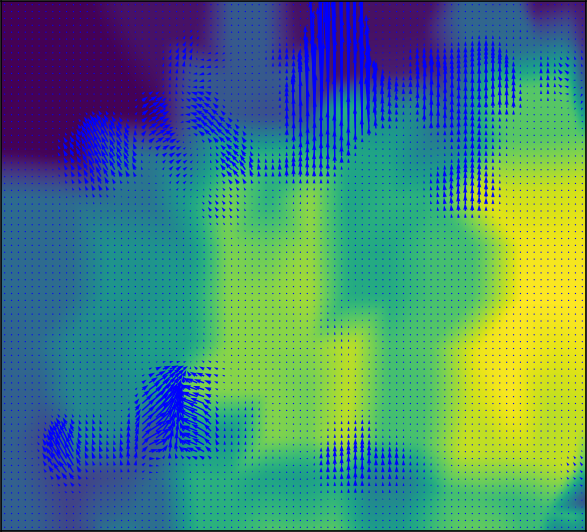}
        \caption{Frame 970}
    \end{subfigure}
    \\
    \begin{subfigure}{0.5\textwidth}
        \centering
        \includegraphics[width=0.45\textwidth]{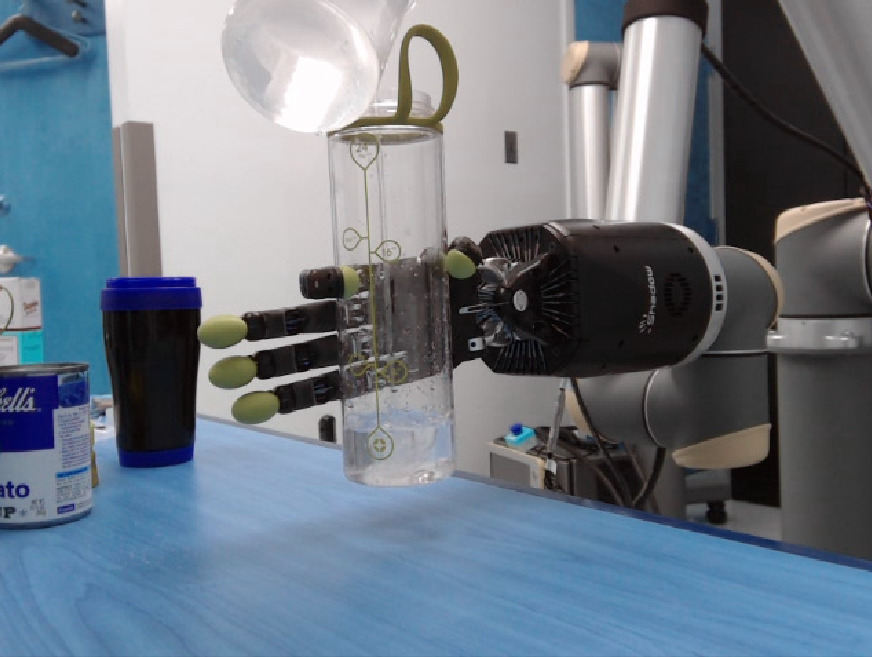}%
        \quad
        \includegraphics[width=0.45\textwidth]{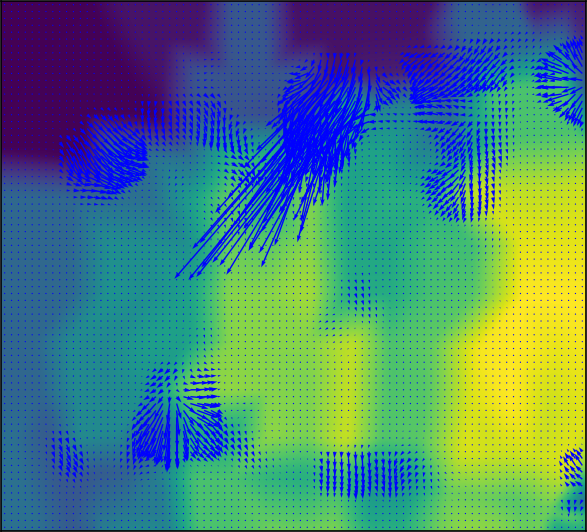}
        \caption{Frame 1500}
    \end{subfigure}
    \\
    \begin{subfigure}{0.5\textwidth}
        \centering
        \includegraphics[width=0.45\textwidth]{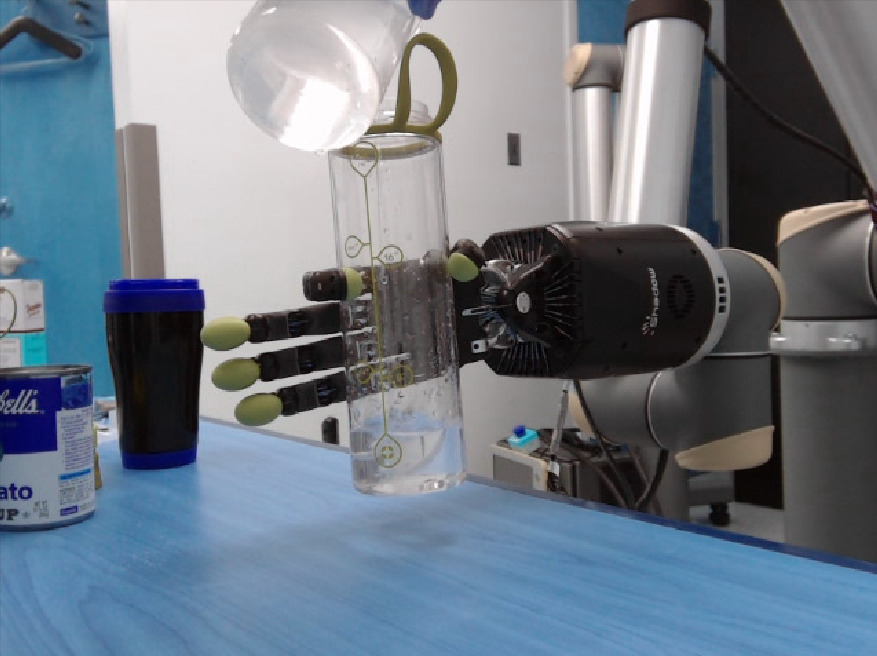}%
        \quad
        \includegraphics[width=0.45\textwidth]{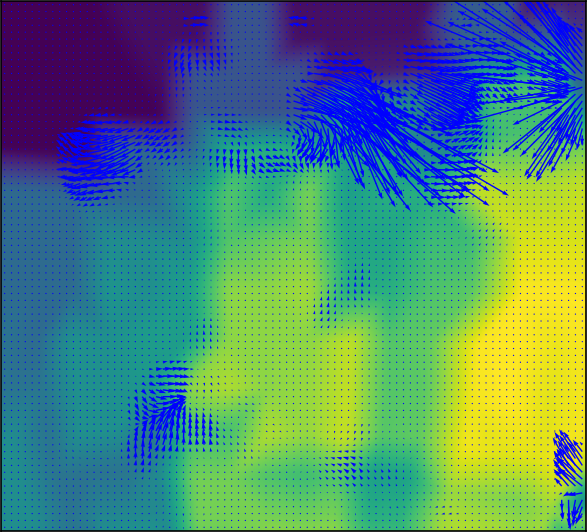}
        \caption{Frame 1750}
    \end{subfigure}
    \\
    \begin{subfigure}{0.5\textwidth}
        \centering
        \includegraphics[width=0.45\textwidth]{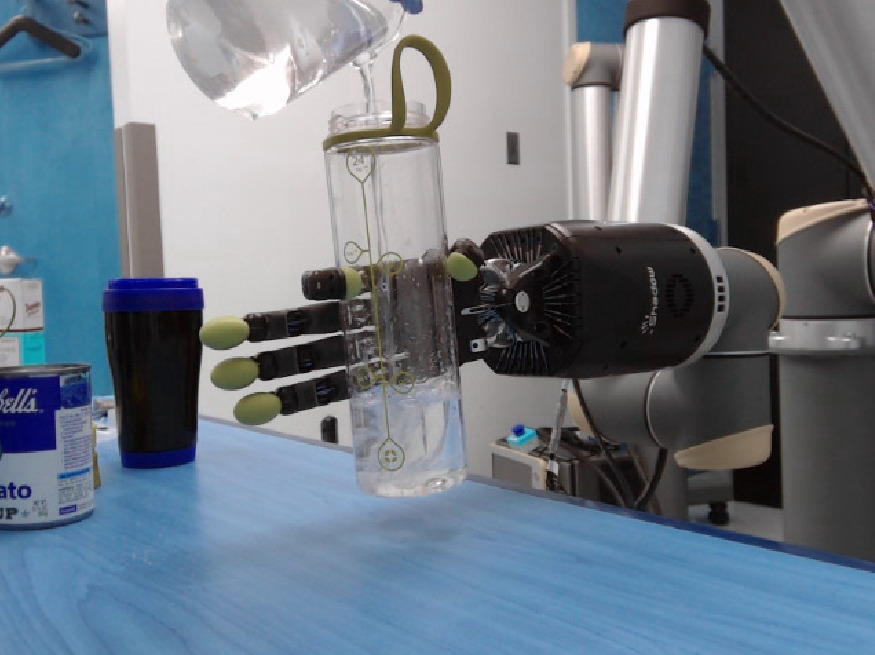}%
        \quad
        \includegraphics[width=0.45\textwidth]{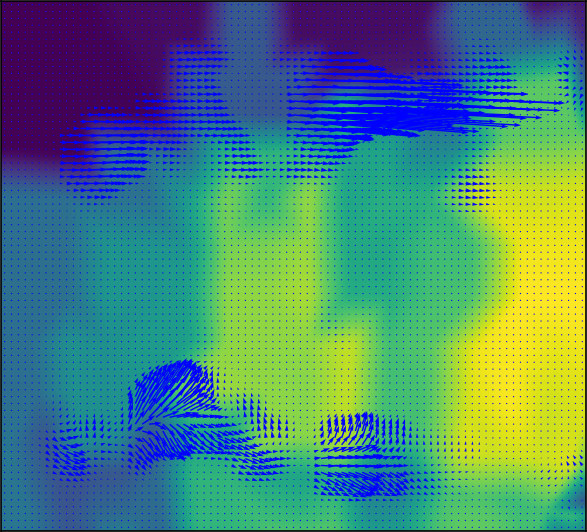}
        \caption{Frame 2250}
    \end{subfigure}
    \\
    \begin{subfigure}{0.5\textwidth}
        \centering
        \includegraphics[width=0.45\textwidth]{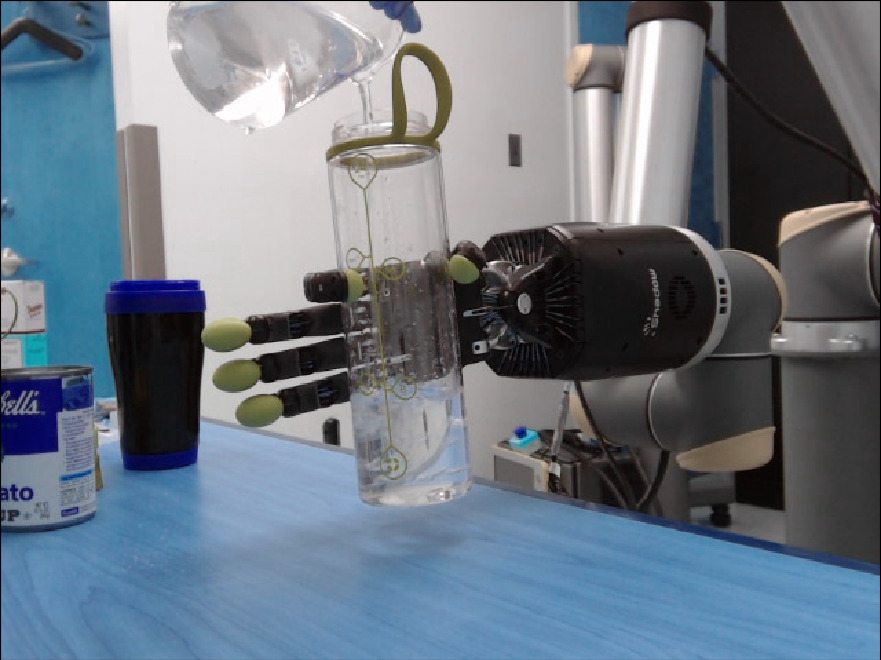}%
        \quad
        \includegraphics[width=0.45\textwidth]{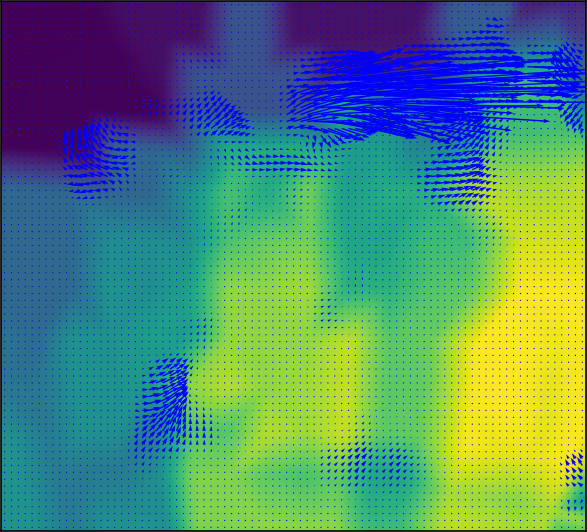}
        \caption{Frame 2390}
    \end{subfigure}
    \caption{Scene configuration and tactile flow sequence for grasping while water is being poured}
    \label{fig:pour_seq}
\end{figure*}
In this part, we discuss our experiments and results on computing \textit{static} tactile flow. These are the flow patterns experienced when the finger and the object in contact is not ``actively'' moving but is held in place by static frictional forces. Such patterns may occur during operating a tool, such as a screwdriver, or when opening a jar for example. In our experiments, we demonstrate a simple instance of the robot grasping an empty bottle into which water is slowly poured from a height. As the weight in the bottle changes, the frictional forces also increase until the fingers are unable to exert sufficient forces to prevent the object from slipping or moving. Note that the robot is completely passive during this operation, i.e. there is no active reactionary force on the bottle to counteract these increasing frictional forces.

The combined taxel and impedance plots are shown in Fig.~\ref{fig:pour_pressures}, where we have used the pressure peaks and troughs to find the regions of interest as before. Each of these segments are color coded to demarcate the important events in the experimentation procedure. The red segment is when the water has just been poured into the bottle, increasing the pressure felt by the fingers. The trough in the green segment is when the pressure is reduced due to inital slippage of the bottle, as it comes loose in the grasp. The magenta and cyan segments represent some rotational motion of the bottle, and some downward motion respectively. Lastly, the yellow segment is when the bottle has started slipping completely.

The corresponding scene configuration and tactile flow sequences are shown in Fig.~\ref{fig:pour_seq}, with the timestamps denoted accordingly.
%%%%%%%%%%%%%%%%%%%%%%%%%%%%%%%%%%%%%%%%%%%%%%%%%%%%%%%%%%%%%%%%%%%%%%%%%%%%%%%%
\section{Conclusions}\label{sec:conclusion}
In this paper, we present the theoretical premise for computing tactile flow using BioTac sensors. We also present a few experiments to validate our claims and methods. However, the practical usefulness of computing tactile flow demands another study and has not been discussed in much detail as part of this paper.

We believe that, like humans, tactile flow is an essential component of robust grasping and proprioception, at par with conventional sensory mechanisms like visual or inertial means that are in use today. In order to achieve general grasping, it is necessary to have an understanding of and appropriate reaction to the forces in play between the object and the fingers. Our method of computational tactile flow provides the foundation for such an understanding, and in future work we aim to present a dynamic grasping mechanism based on feedback from these tactile flow data.

Our plan for upcoming research is to use this computed tactile flow in order to facilitate better, more robust grasping strategies. We hope to use tactile flow as feedback in a grasping pipeline, in order to obtain dynamic and robust grasps, similar to how humans perform grasping without constant visual feedback.

Also, we consider tactile flow patterns to be representative of motion types in a given task. Thus, flow patterns can provide a way to encode motion primitives for a given task, and that can be emulated by the robot when trying to replicate a particular task. This gives rise to a general learning paradigm for grasping of novel objects, based on task goals. Latest machine learning paradigms combined with conventional control algorithms can prove to be highly effective in learning these flow patterns, and reproducing them with the added benefit of generalizing to various scenarios.

%%%%%%%%%%%%%%%%%%%%%%%%%%%%%%%%%%%%%%%%%%%%%%%%%%%%%%%%%%%%%%%%%%%%%%%%%%%%%%%%
\bibliographystyle{plain}
\bibliography{tactile_flow}
% \nocite{*}
\end{document}